\documentclass[10pt,twocolumn,letterpaper]{article}

\usepackage[pagenumbers]{cvpr}
\usepackage{times}
\usepackage{epsfig}
\usepackage{graphicx}
\usepackage{amsmath}
\usepackage{amssymb}

\usepackage{booktabs}
\usepackage{multirow}
\usepackage{float}
\usepackage{amsthm,amsmath,amssymb}
\usepackage{mathrsfs}
\usepackage{bm}
\usepackage{bbding}
\usepackage{float}
\usepackage[normalem]{ulem}
\useunder{\uline}{\ul}{}
\usepackage[table,xcdraw]{xcolor}

\usepackage{subcaption}

\usepackage[pagebackref=true,breaklinks=true,letterpaper=true,colorlinks,bookmarks=false]{hyperref}

\usepackage[capitalize]{cleveref}
\crefname{section}{Sec.}{Secs.}
\Crefname{section}{Section}{Sections}
\Crefname{table}{Table}{Tables}
\crefname{table}{Tab.}{Tabs.}

\begin{document}

\title{High-Quality Real-Time Rendering \\ Using Subpixel Sampling Reconstruction}

\author{
Boyu Zhang$^{1,2}$, Hongliang Yuan$^{1}$, Mingyan Zhu$^{1,3}$, Ligang Liu$^4$, Jue Wang$^1$\\
$^1$Tencent AI Lab, $^2$Southeast University, $^3$Tsinghua University,\\ $^4$University of Science and Technology of China\\
{\tt\small byz@seu.edu.cn, 11488336@qq.com, zmy20@mails.tsinghua.edu.cn}\\
{\tt\small lgliu@ustc.edu.cn, maxjwang@tencent.com}}


\vspace{-0.200cm}
\twocolumn[{%
\maketitle

\vspace{-0.75cm}

\begin{figure}[H]
\captionsetup[subfigure]{justification=centering}
\hsize=\textwidth 
\centering
\begin{subfigure}{0.195\textwidth}
    \includegraphics[width=\textwidth]{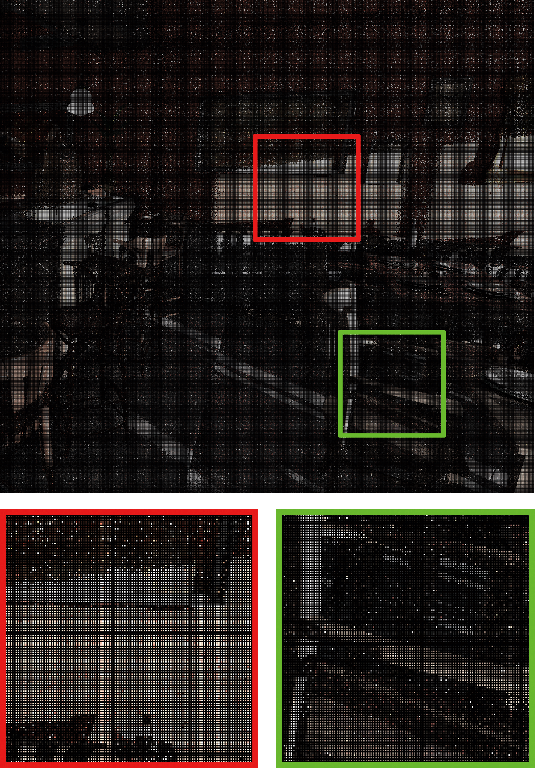}
    \caption{1/4-spp input\\ \ }
\end{subfigure}
\begin{subfigure}{0.195\textwidth}
    \includegraphics[width=\textwidth]{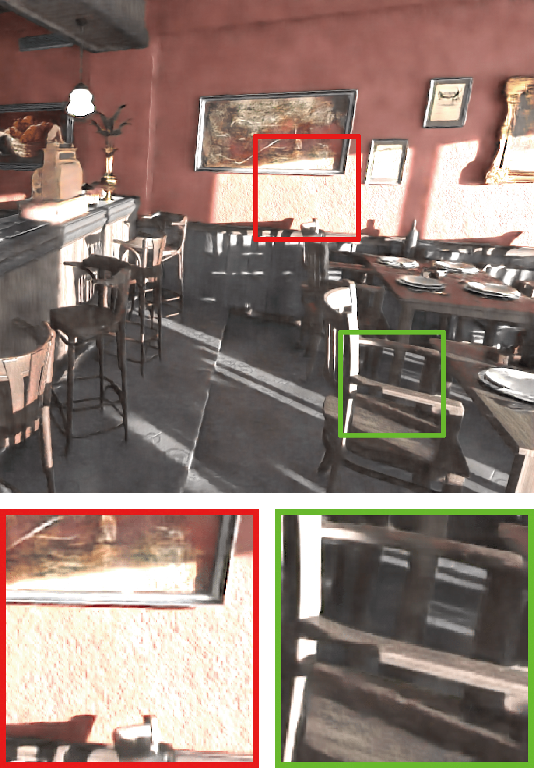}
    \caption{NSRR\cite{Xiao2020} \\ FPS=89, SSIM=0.7737}
\end{subfigure}
\begin{subfigure}{0.195\textwidth}
    \includegraphics[width=\textwidth]{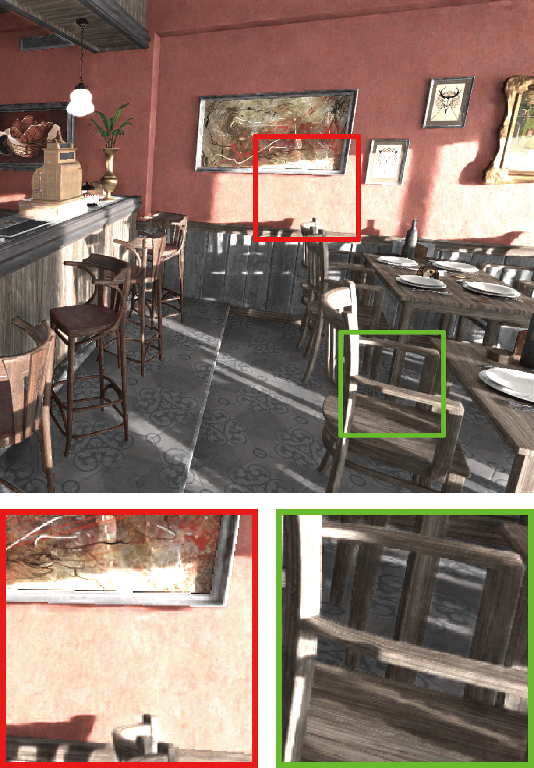}
    \caption{RAE\cite{Chaitanya17} \\ FPS=96, SSIM=0.7556}
\end{subfigure}
\begin{subfigure}{0.195\textwidth}
    \includegraphics[width=\textwidth]{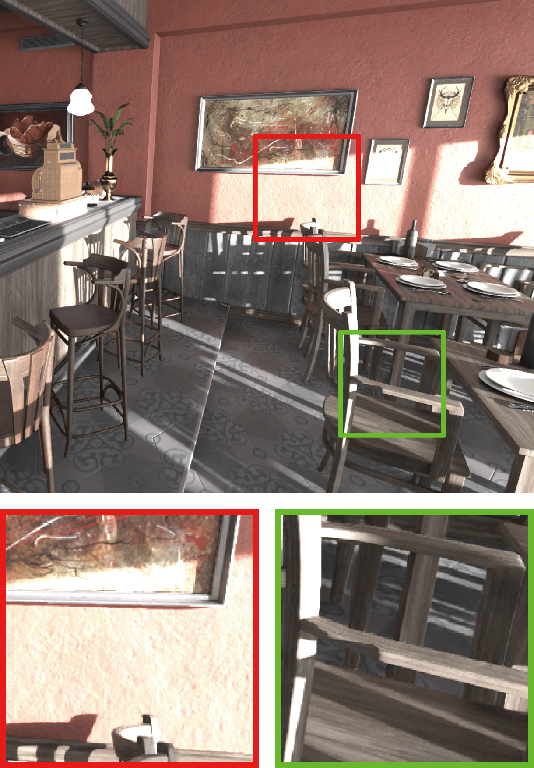}
    \caption{Ours \\ FPS=131, SSIM=0.9036}
\end{subfigure}
\begin{subfigure}{0.195\textwidth}
    \includegraphics[width=\textwidth]{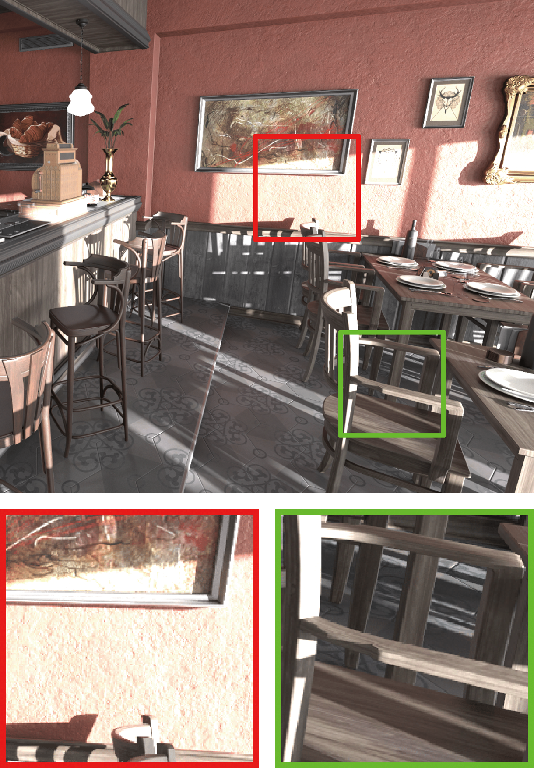}
    \caption{Ground truth\\ \ }
\end{subfigure}

  \caption{We present a novel Monte Carlo sampling strategy (a) that uses less than one sample per pixel (spp) to accelerate high-resolution rendering and enable real-time performance. To complement this strategy, we developed a denoiser (d) that produces high-quality reconstructed images with improved details and faster inference time than previous state-of-the-art rendering denoisers (b) (c). The ground truth image is rendered at 32768-spp (e) for reference.
}
  \label{fig:teaser}
\end{figure}
}]

\begin{abstract}
\vspace{-0.40cm}
Generating high-quality, realistic rendering images for real-time applications generally requires tracing a few samples-per-pixel (spp) and using deep learning-based approaches to denoise the resulting low-spp images. Existing denoising methods have yet to achieve real-time performance at high resolutions due to the physically-based sampling and network inference time costs. In this paper, we propose a novel Monte Carlo sampling strategy to accelerate the sampling process and a corresponding denoiser, subpixel sampling reconstruction (SSR), to obtain high-quality images. Extensive experiments demonstrate that our method significantly outperforms previous approaches in denoising quality and reduces overall time costs, enabling real-time rendering capabilities at 2K resolution.

\end{abstract}

\vspace{-0.35cm}


\section{Introduction}

Rendering realistic images for virtual worlds is a key objective in many computer vision and graphics tasks \cite{Wimbauer_2022_CVPR, Petersen_2022_CVPR, Li_2022_CVPR, huo2021survey, Wang_2021_ICCV, Cole_2021_ICCV}, with applications in animation production\cite{dahlberg2019machine}, VR/AR world generation\cite{2018vr2}, and virtual dataset synthesis \cite{ge2022neural} \etc. One widely used technique for this purpose is Monte Carlo (MC) sampling \cite{seila1982simulation}, which is highly versatile but typically requires a large number of samples to achieve accurate results. Despite continuously increasing computational power, the time required for realistic rendering remains a practical constraint, with high-quality images often taking hours to generate. Using low samples-per-pixel (spp) can speed up this process, but lead to visually distracting noise. To mitigate this issue, post-processing techniques have been developed, known as MC denoising, which normally have lower time costs than physically-based renderer and are widely used in modern game engines\cite{Chaitanya17, dlss, Xiao2020}.

Most existing MC denoising methods \cite{Edelsten2019, Xiao2020, Chaitanya17, Mustafa21, Meng20, Hasselgren2020, Fan2021} employ deep learning-based approaches to remove noise from images generated with more than 1-spp. While \cite{Chaitanya17, Meng20, Fan2021} attempt to develop methods to accelerate the overall process by working with low-sample data, they have yet to achieve real-time frame rates at high resolutions, as 1-spp remains time-consuming. Other methods \cite{Edelsten2019, Xiao2020, Mustafa21} focus on designing more efficient post-processing modules in the image space to handle noisy images, but they tend to produce aliased rendered pixels at low-sample images. Additionally, the complex network structures of these methods impose heavy burdens on inference time.

To achieve real-time performance of generating high-quality, high-resolution realistic images, we present a novel MC sampling strategy, subpixel sampling, for reducing the computational cost of physically-based rendering. Furthermore, we propose a denoising method subpixel sampling reconstruction (SSR), which is tailored to this sampling strategy. 

\noindent\textbf{Subpixel sampling.}
Subpixel sampling strategy generates images with less than 1-spp. To obtain this, we divide the frame at the target resolution into consecutive, non-overlapping tiles with size $2\times2$ and then compute only one ray-traced pixel per tile (we refer to it as 1/4-spp). This strategy allows us to use these reliable samples to interpolate the missing pixels with the GBuffers\cite{gbuffer} at the target resolution, which are available nearly for free. We developed a hybrid ray tracer based on Vulkan\cite{sellers2016vulkan} to export datasets. By utilizing subpixel sampling, the cost of rendering time can be reduced by a third.

\noindent\textbf{Subpixel sampling reconstruction.}
 Our SSR contains two parts: a temporal feature accumulator and a reconstruction network. The former warps previous frames to align with the current frame at the target resolution and accumulates subpixel samples and GBuffers from the previous frame based on the temporal accumulation factor, which is computed according to the correlation of the current and previous frame, effectively expanding the perception field of pixels. Once subpixel samples are collected, we move on to the second component, our reconstruction network. This is a multi-scale U-Net \cite{UNet2015} with skip connections, which enables us to reconstruct the desired high-resolution image.


The key points of our contribution can be summarized as
follows:

\begin{itemize}
\item We propose a new Monte Carlo sampling strategy called subpixel sampling, which significantly reduces the sampling time of physically-based rendering to 1/3.

\item We introduce a denoising network, SSR, to reconstruct high-quality image sequences at real-time frame rates from rendering results using subpixel sampling strategy.

\item Our model yields superior results compared to existing state-of-the-art approaches. Moreover, we are the first to achieve real-time reconstruction performance of 2K resolution with 130 FPS. 

\item A realistic synthesised dataset is built through our sparse sampling ray tracer. We will release the ray tracer and dataset for research purpose.
\end{itemize}

\section{Related Work}

\subsection{Monte Carlo Denoising}

Monte Carlo (MC) Denoising is widely applied in rendering realistic images. Traditional best-performing MC denoisers were mainly based on local neighborhood regression models \cite{Zwicker2015}, includes zero-order regression \cite{Rousselle2012, Delbracio2014, Li2012, Kalantari2015, Rousselle2013, Moon2013}, first-order regression \cite{Bauszat2011, Bitterli2016, Moon2014} and even higher-order regression models \cite{Moon2016}.The filtering-based methods are based on using the auxiliary feature buffers to guide the construction of image-space filters. Most of the above methods run in offline rendering. To increase the effective sample count, real-time denoisers leverage temporal accumulation between frames over time to amortize supersampling \cite{Yang2009}, \ie temporal anti-aliasing (TAA). The previous frame is reprojected according to the motion vector and blended with the current frame using a temporal accumulation factor, which can be constant \cite{Schied2017, Mara2017, Meng20} or changed \cite{Schied2018} across different frames. The fixed temporal accumulation factor inevitably leads to ghosting and temporal lag. By setting the parameter adaptively, the temporal filter can fastly respond to times in case of sudden changes between frames. Yang et al. \cite{Yang2020} survey recent TAA techniques and provide an in-depth analysis of the image quality trade-offs with these heuristics. Koskela et al. \cite{Koskela2019} propose a blockwise regression for real-time path tracing reconstruction and also do accumulation to improve temporal stability.

\subsection{Deep Learning-Based Denoising}

Recently, with the advent of powerful modern GPUs, many works utilize
CNN to build MC denoisers. \cite{Bako2017, Vogels18} use deep CNN to estimate the local per-pixel filtering kernels used to compute each denoised pixel from its neighbors. Dahlberg et al. \cite{Dahlberg2019} implement the approach of \cite{ Vogels18} as a practical production tool used on the animated feature film. Layer-based denoiser \cite{Munkberg2020} designs a hierarchical kernel prediction for multi-resolution denoising and reconstruction.  Since the high computational cost of predicting large filtering kernels, these methods mostly target offline renderings. There are also other methods \cite{Kuznetsov2018, Xu2019, Gharbi2019, Yu2021, back2022self} that target denoising rendering results at more than 4 spp. To reduce the overhead of kernel prediction, Fan et al. \cite{Fan2021} predict an encoding of the kernel map, followed by a high-efficiency decoder to construct the complete kernel map. Chaitanya et al. \cite{Chaitanya17} proposed a recurrent connection based on U-Net \cite{UNet2015} to improve temporal stability for sequences of sparsely sampled input images. Hasselgren et al. \cite{Hasselgren2020} proposed a neural spatio-temporal joint optimization of adaptive sampling and denoising with a recurrent feedback loop. Hofmann et al. \cite{Hofmann2021} also utilized the neural temporal adaptive sampling architecture to denoise rendering results with participating media. Xiao et al. \cite{Xiao2020} presented a neural supersampling method for TAA, which is similar to deep-learned supersampling (DLSS) \cite{Edelsten2019}. Meng et al. \cite{Meng20} denoised 1-spp noisy input images with a neural bilateral grid at real-time frame rates. Mustafa et al. \cite{Mustafa21} adopted dilated spatial kernels to filter the noisy image guiding by pairwise affinity over the features. Compare with these denoising framework targeting for more than 1-spp renderings, our method is designed to work with 1/4-spp for cutting off rendering time cost.

\section{Method}

\begin{figure}[t]
  \centering
\begin{subfigure}{0.95\linewidth}
\includegraphics[width=\textwidth]{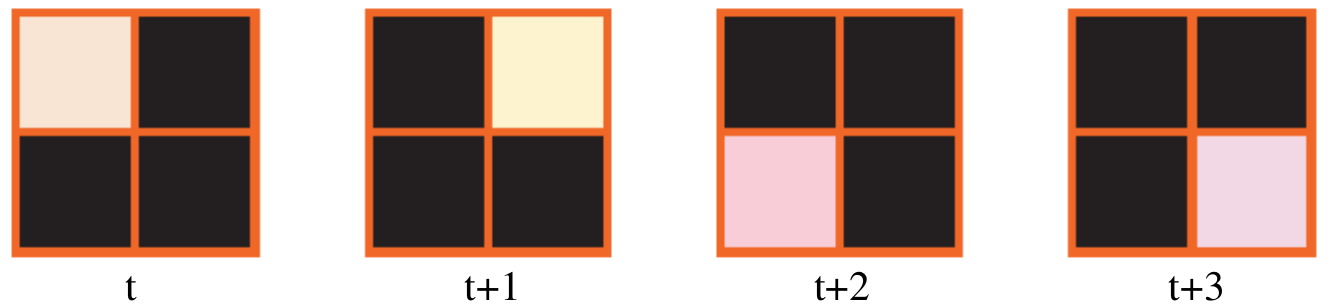}
\caption{Subpixel sampling strategy}
\label{fig:sampling-a}
\end{subfigure}
\begin{subfigure}{0.48\linewidth}
\includegraphics[width=\textwidth]{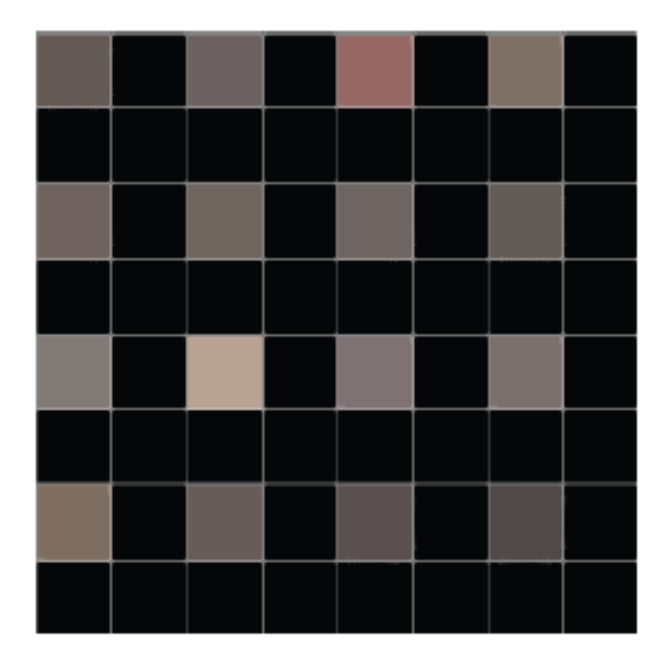}
\caption{Pixels}
\label{fig:sampling-b}
\end{subfigure}
\begin{subfigure}{0.48\linewidth}
\includegraphics[width=\textwidth]{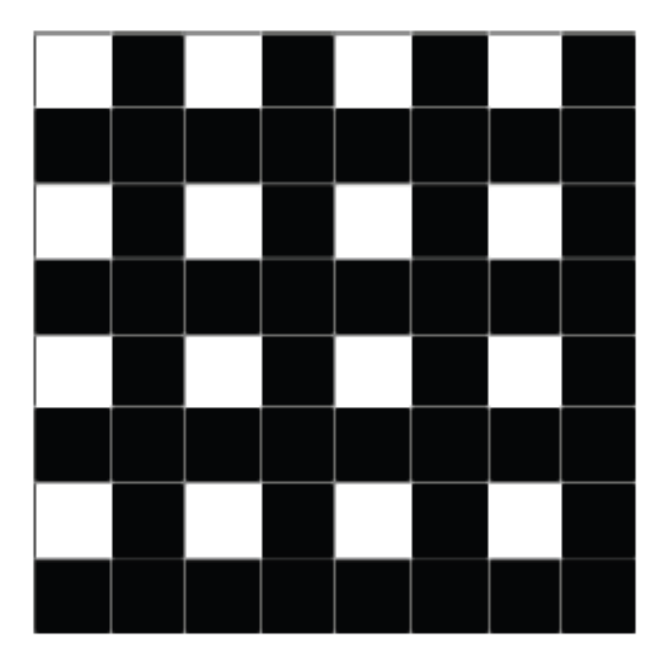}
\caption{Mask map}
\label{fig:sampling-c}
\end{subfigure}

  \caption{(a) Sampling of a $2\times2$ tile from consecutive four frames. The sampled and unsampled pixels are drawn in color and in black (with value 0), respectively. 
  (b) Pixels of a sub-patch example in a rendered image.
  (c) The corresponding mask map of patch (b) is depicted in white pixels with a value of 1, while black pixels indicate a value of 0.
}
\label{fig:sparse_sampling}
\end{figure}

\subsection{Subpixel Sampling}

To address the high computational cost of rendering when the samples-per-pixel (spp) exceeds 1, we develop subpixel sampling strategy that enables us to generate images with spp less than 1.

\noindent\textbf{1/4-spp pattern}
Our strategy involves dividing each frame into non-overlapping $2\times2$ tiles and applying a Monte Carlo-based ray tracing method to solve the rendering equation \cite{Kajiya1986} for one pixel in each tile. We term this process 1/4-spp pattern. To maintain data balance, we shift the sampling position to ensure that each pixel is sampled in the consecutive four frames at time steps $t$ to $t+3$, as illustrated in \cref{fig:sampling-a}.

\begin{figure}[t]
\begin{center}
\begin{subfigure}{0.24\linewidth}
\includegraphics[width=\textwidth]{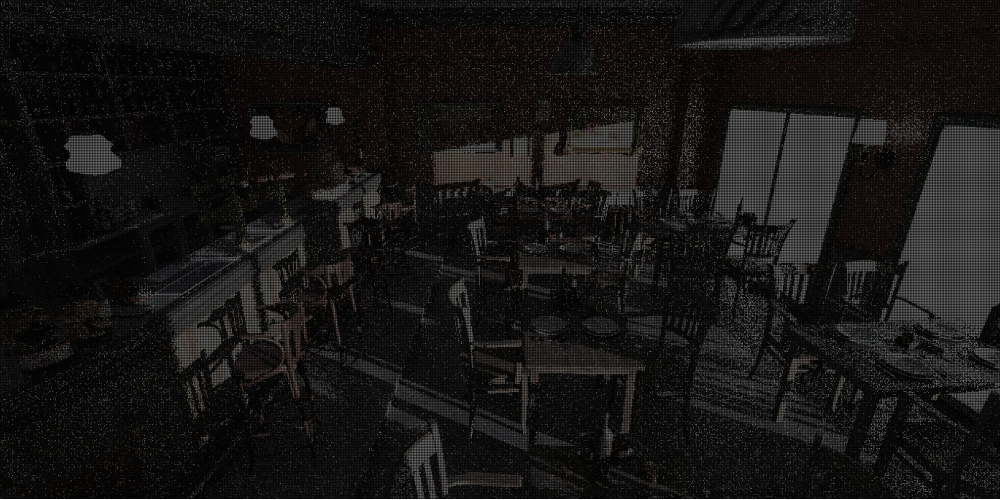}
\caption{RGB Color}
\label{gb1}
\end{subfigure}
\begin{subfigure}{0.24\linewidth}
    \includegraphics[width=\textwidth]{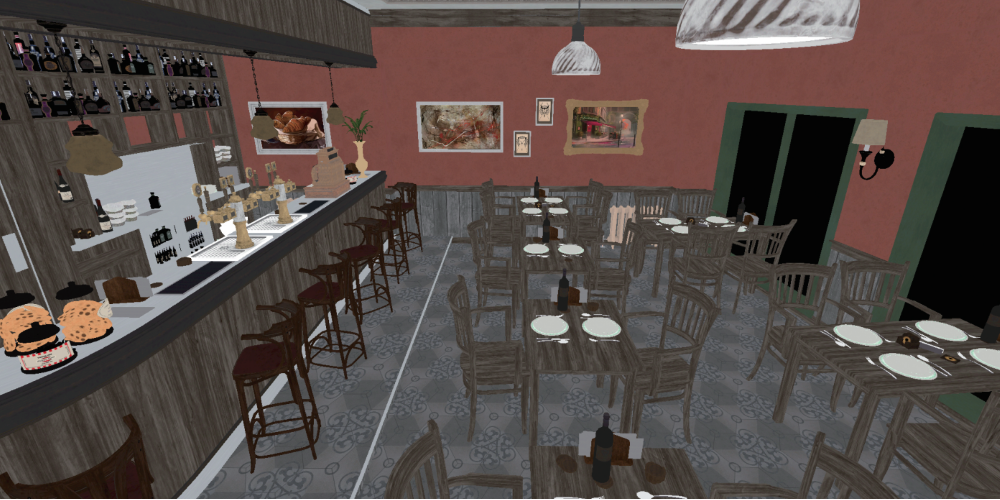}
    \caption{Albedo}
\label{gb2}
\end{subfigure}
\begin{subfigure}{0.24\linewidth}
    \includegraphics[width=\textwidth]{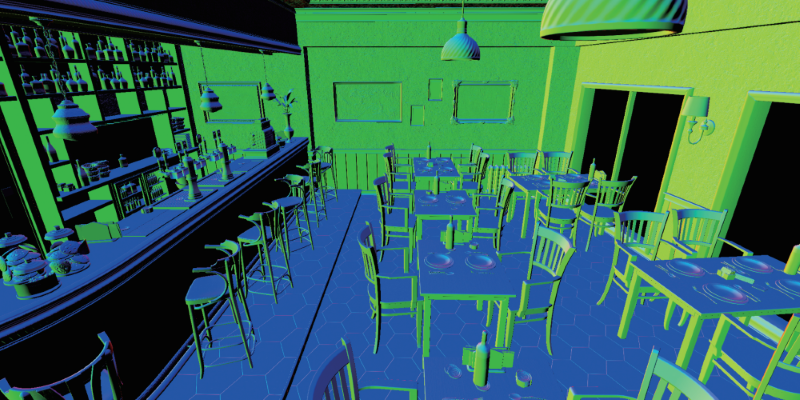}
    \caption{Normal}
\label{gb3}
\end{subfigure}
\begin{subfigure}{0.24\linewidth}
    \includegraphics[width=\textwidth]{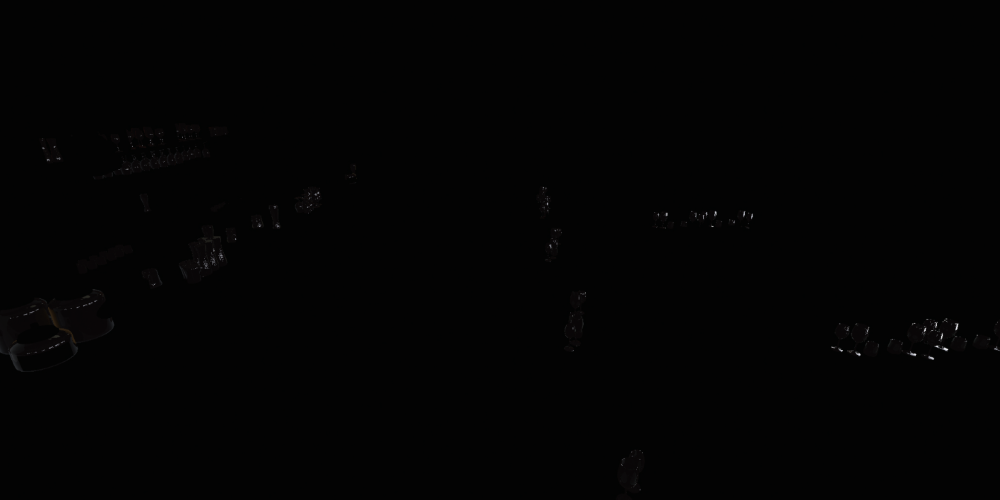}
    \caption{Transparent}
\label{gb4}
\end{subfigure}
\begin{subfigure}{0.24\linewidth}
    \includegraphics[width=\textwidth]{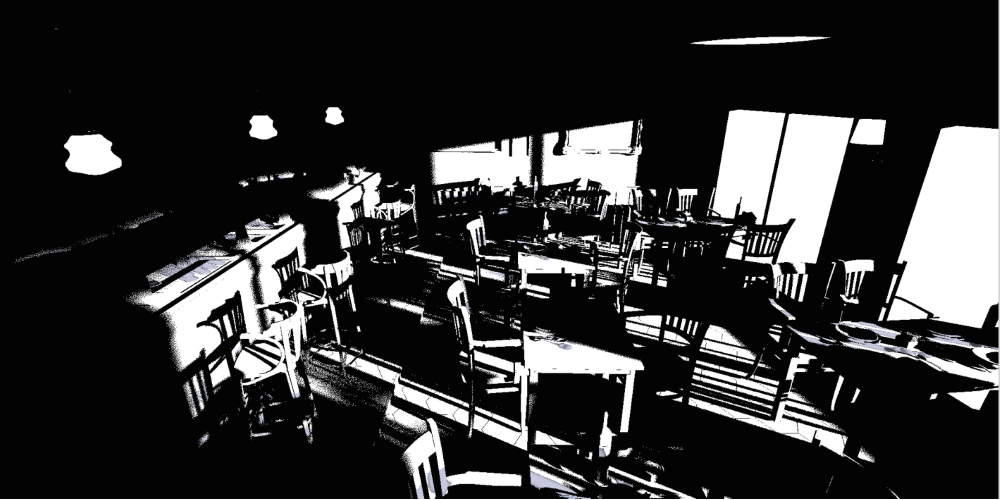}
    \caption{Shadow}
\label{gb5}
\end{subfigure}
\begin{subfigure}{0.24\linewidth}
    \includegraphics[width=\textwidth]{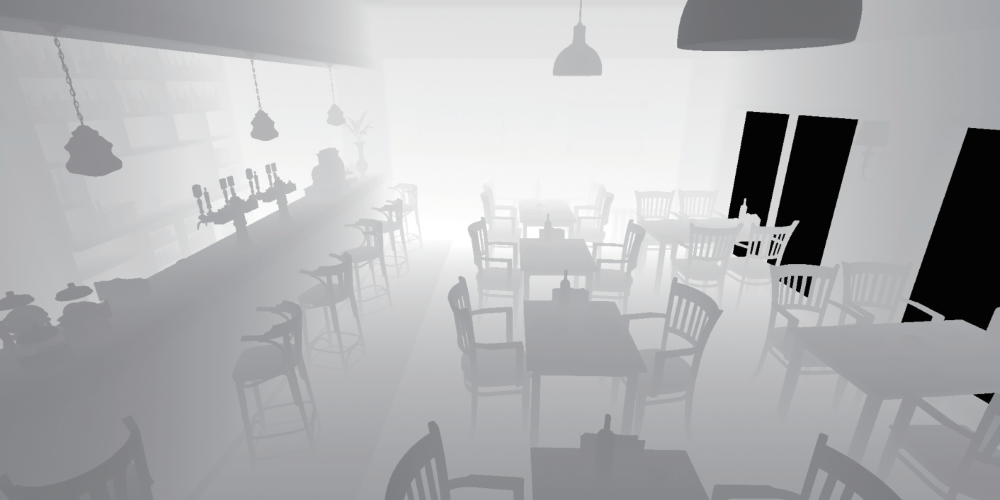}
    \caption{Depth}
\label{gb6}
\end{subfigure}
\begin{subfigure}{0.24\linewidth}
    \includegraphics[width=\textwidth]{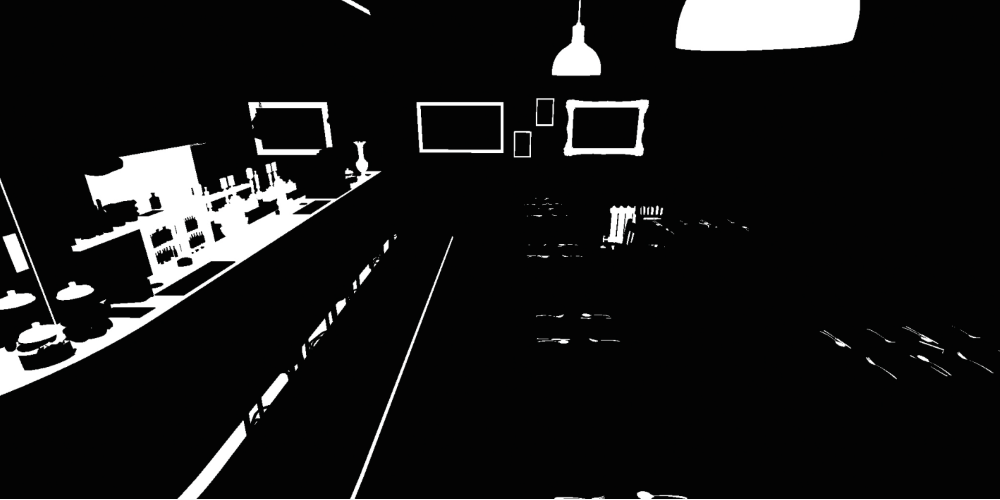}
    \caption{Metallic}
\label{gb7}
\end{subfigure}
\begin{subfigure}{0.24\linewidth}
    \includegraphics[width=\textwidth]{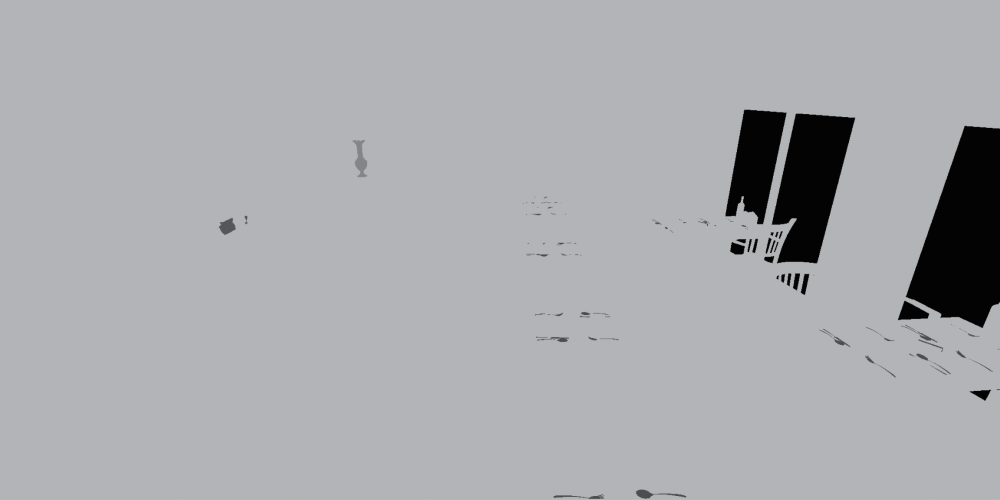}
    \caption{Roughness}
\label{gb8}
\end{subfigure}
\end{center}
\caption{Dumped buffers from our ray tracer.}
\label{fig:gbuffer}
\end{figure}

\noindent\textbf{GBuffers}
We leverage rasterization pipeline to efficiently output high-resolution GBuffers. In detail, for each frame, we dump 1/4-spp RGB color $\textbf{c}\in \mathbb{R}^3$ (\cref{gb1}) and features $\textbf{f}\in \mathbb{R}^{15}$. These features comprise four 3D vectors (albedo, normal, shadow, and transparent) and three 1D vectors (depth, metallic, roughness), as shown in \cref{gb2,gb3,gb4,gb5,gb6,gb7,gb8}.

\noindent\textbf{Mask map}
As the sampled subpixels are ray traced at the high resolution, their RGB values are reliable for the target resolution. We propose generating an additional mask map, where the sampled positions have a value of 1 and unsampled positions to be 0, to denote reliable pixels, as shown in \cref{fig:sampling-c}. This mask map, performed as a confidence map, is expected to guide our temporal feature accumulator (Section \cref{sec:temporal_accumulator}) to predict reasonable weight. To this end, we incorporate the mask map into the GBuffers.

\noindent\textbf{Demodulation}
Similar to the previous approach \cite{Chaitanya17}, we utilize the albedo (or base color) to demodulate the RGB image. Next, the resulting untextured irradiance $x$ is transformed into log space using the natural logarithm function, i.e., $\ln(1+x)$. However, our method differs in that once the untextured irradiance has been reconstructed, we re-modulate it using the accumulated albedo predicted by our temporal feature accumulator.

\subsection{Subpixel Sampling Reconstruction}

\begin{figure*}[t]
\begin{center}
 \includegraphics[width=\textwidth]{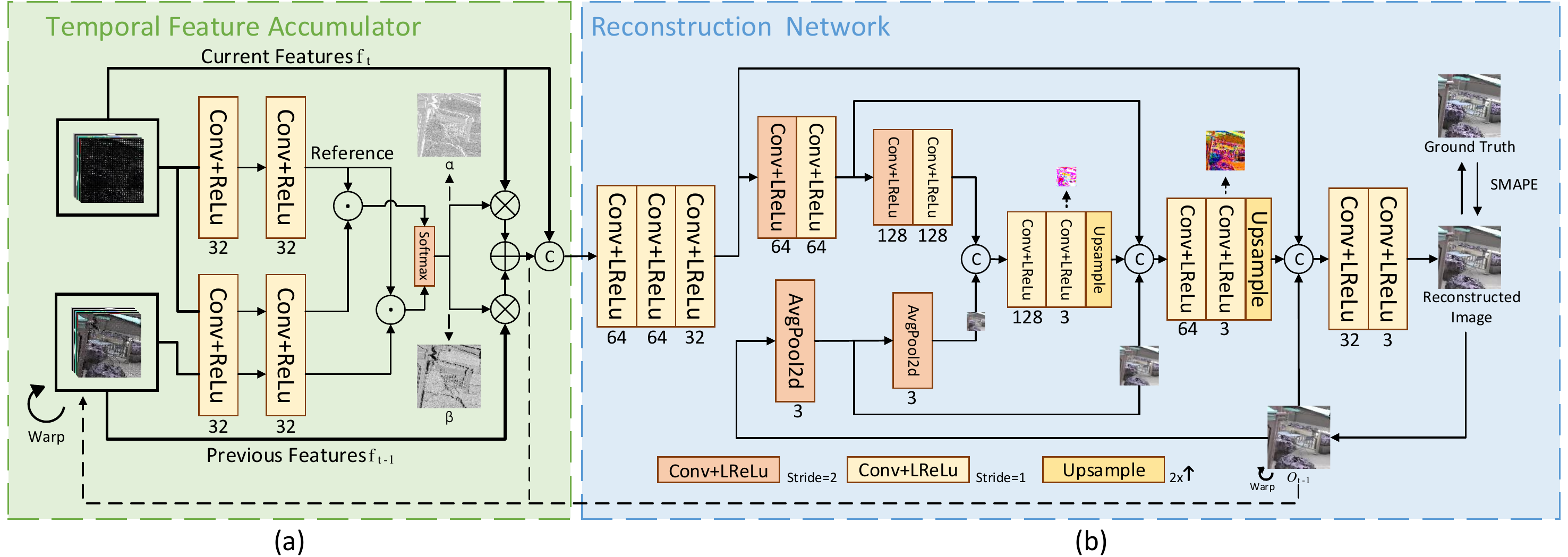}
\end{center}

\caption{Subpixel sampling reconstruction consists of two modules: the temporal feature accumulator (left) and the reconstruction network (right) . The numbers under each network layer represent the output channels at corresponding layers. The operator $\odot$ denotes dot product between features. {\small{\textcircled{c}}} indicates concatenation operation. $\oplus$ and $\otimes$ represent element-wise addition and multiplication, respectively. Note that all frames shown here are demodulated by albedo.}
\label{fig:net_structure}
\end{figure*}

We designed subpixel sampling reconstruction (SSR) to recover temporally stable video from 1/4-spp image sequences at real-time frame rates. \cref{fig:net_structure} shows the detailed architecture of SSR, which comprises two modules: the temporal feature accumulator (in green) and the reconstruction network (in blue).

\subsubsection{Temporal Feature Accumulator}
\label{sec:temporal_accumulator}

\begin{figure}[t]
\begin{center}
 \includegraphics[width=0.42\textwidth]{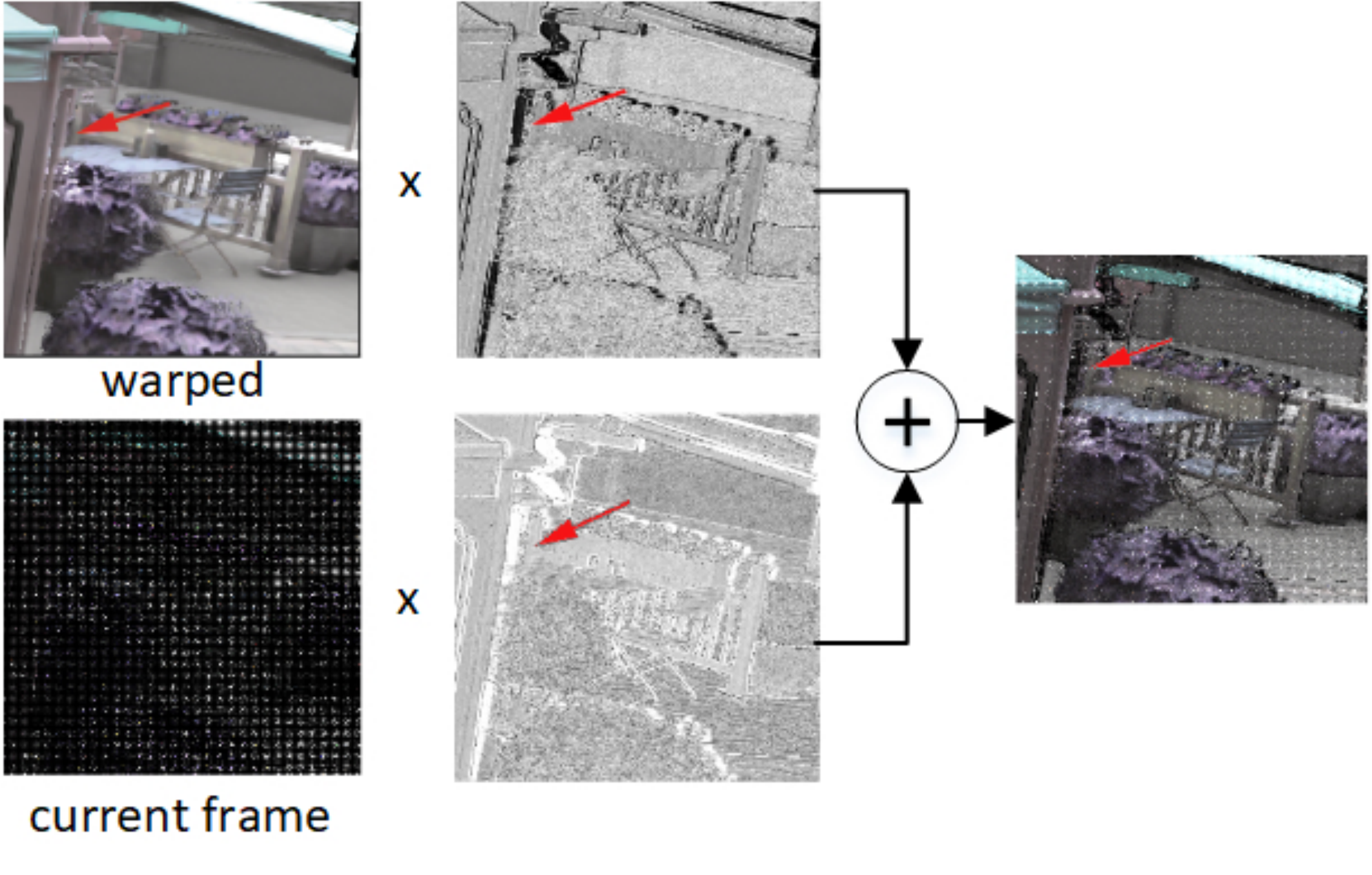}
\end{center}
\caption{ Illustration of accumulating untextured irradiance by our temporal feature accumulator Warped irradiance by motion vector has ghosting artifacts (red arrow), which can be removed by giving lower weight in these areas.}
\label{fig:accumulator}
\end{figure}

The temporal feature accumulator module consists of two neural networks,  each with two convolution layers that have a spatial support of $3 \times 3$ pixels. One network accepts all features and mask of current frame as input and outputs reference embedding. The other computes embeddings for the current features $\bm{f}_t$ and warped previous features $\bm{f}_{t-1}$. These two embeddings are then pixel-wise multiplied to the reference embedding and then through softmax($\cdot$) to get $\mathbf{\alpha}$ and $\mathbf{\beta}$ ($\mathbf{\alpha}+\mathbf{\beta}=1$) blending factors for current features and previous features, respectively. 

All features in \cref{fig:gbuffer} are accumulated through above process. Take untextured irradiance as an example, as illustrated in \cref{fig:accumulator}, we use the following equation to accumulate untextured irradiance $\textbf{e}$ over the frame:

\begin{equation}
\textbf{e}^a_{t} = \mathbf{\alpha} \mathcal{W}(\textbf{e}^a_{t-1}) + \mathbf{\beta} \textbf{e}_{t},
\label{equation:acc}
\end{equation}

where $\textbf{e}^a_{t}$ is accumulated irradiance until $t$ frame, $\textbf{e}_{t}$ is irradiance for $t$ frame. For the first frame, we set $\textbf{e}^a_{t-1}$ to $\textbf{e}_{t}$. $\mathcal{W}(\cdot)$ is a warping operator that reprojects previous frame to current one using motion vector. 

The temporal feature accumulator serves as a vital role for producing temporally stable results. Firstly, it can detect and remove disoccluded pixels and ghosting artifacts that traditional motion vectors cannot handle accurately. Secondly, since our input images are sparsely sampled, this module helps gather more finely sampled pixels over time.


\subsubsection{Reconstruction Network}
\label{sec:reconstruction}

Our reconstruction network extends U-Net \cite{UNet2015} with skip connections \cite{Mao2016}. In contrast to other U-Net based denoising methods \cite{Chaitanya17}, our approach predicts two coarse-scale images at the first two decoder stages, rather than predicting dense features at these stages. This modification not only leads to faster inference but also results in high-quality images with superior quantitative metrics (see \cref{sec:ab}).

To generate a high-quality image for the current frame, we concatenate the current and accumulated features and feed them into our reconstruction network. Additionally, we input the warped denoised image from the previous frame, which enhances the temporal stability of image sequences (see \cref{sec:ab}). The reconstruction network consists of three encoder layers that produce three scale features.

Retaining temporal feedback at multiple scales is also a crucial step. To accomplish this, we downsample the warped denoised image from the previous frame using pool with a stride of two and pass it to each encoding stage. At the decoder stage, we concatenate the features and the warped denoised image at the same scale and feed them into a tile with two convolution layers. At the first two decoder stages, the image in RGB-space is produced and upsampled. This upsampled image is then passed to the next decoder stage. The multi-scale feedback enables us to achieve a sufficiently large temporal receptive field and efficiently generate high-quality, temporally stable results.

\subsection{Loss}
\label{Losses}

We use the symmetric mean absolute percentage error (SMAPE):
\begin{equation}
    \ell(\textbf{r}, \textbf{d}) = \frac{1}{3N}  \sum_{p=1}^{p=N} \sum_{c=1}^{c=3} \frac{\left| \textbf{d}_{p,c} - \textbf{r}_{p,c} \right|}{\left| \textbf{d}_{p,c} \right| + \left| \textbf{r}_{p,c} \right| + \varepsilon },
\end{equation}
where $N$ is the number of pixels and $\varepsilon$ is a tiny perturbation, $\textbf{d}$ and $\textbf{r}$ are the denoised frame and the corresponding reference frame, respectively. 

Our loss combines two parts, the first one is computed on a sequence of 5 continuous frames, including spatial loss $\ell_s=\ell(\textbf{r}, \textbf{d})$, temporal loss $\ell_t=\ell(\Delta\textbf{r}, \Delta\textbf{d})$ where $\Delta$ is temporal gradient computed between two consecutive frames, relative edge loss $\ell_e=L_1(\frac{\nabla\textbf{d}}{\textbf{r} + \varepsilon}, \frac{\nabla\textbf{r}}{\textbf{r} + \varepsilon})$, where gradient $\nabla$ is computed using a High Frequency Error Norm (HFEN), an image comparison metric from medical imaging \cite{Ravishankar2011}. As suggested by Chaitanya et al. \cite{Chaitanya17}, we assign higher weight to three loss functions ($\ell_s$, $\ell_t$ and $\ell_e$) of frames later in the sequence to amplify temporal gradients. For our training sequence of 5 images, we use (0.05, 0.25, 0.5, 0.75, 1). The second part is warped temporal loss $\ell_{wt}=\ell(\omega\textbf{r}, \omega\textbf{d})$ where $\omega\textbf{r}=r_4-\mathcal{W}(r_3)$, $\mathcal{W}(\cdot)$ is a warping operator that reprojects previous frame to current one. We also include albedo loss $\ell_a=\ell(\textbf{a}_{acc}, \textbf{a}_r)$. $\textbf{a}_{acc}$ is accumulated albedo computed by our temporal feature accumulator network. We only compute albedo loss on last frame and warped temporal loss on last two frames.

We use a weighted combination of these losses as the overall loss:
\begin{equation}
    \ell = \lambda_s\ell_s + \lambda_t\ell_t + \lambda_e\ell_e + \lambda_w\ell_{wt} + \lambda_a\ell_a.
\end{equation}

\section{Experiments}

\subsection{Datasets and Metrics}
\label{dataset}
\begin{figure}[t]
\begin{subfigure}{0.32\linewidth}
    \includegraphics[width=\textwidth]{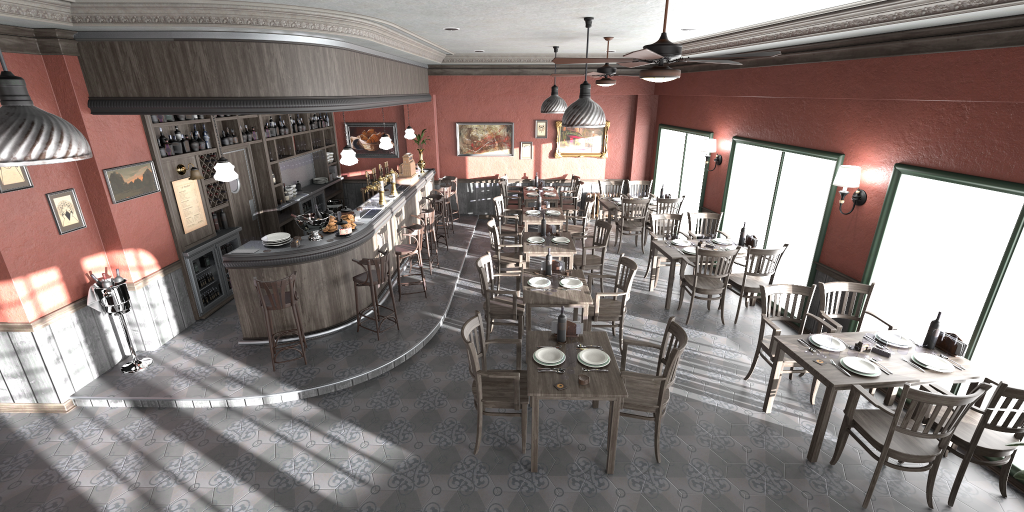}
    \caption{BistroInterior}
\end{subfigure}
\begin{subfigure}{0.32\linewidth}
    \includegraphics[width=\textwidth]{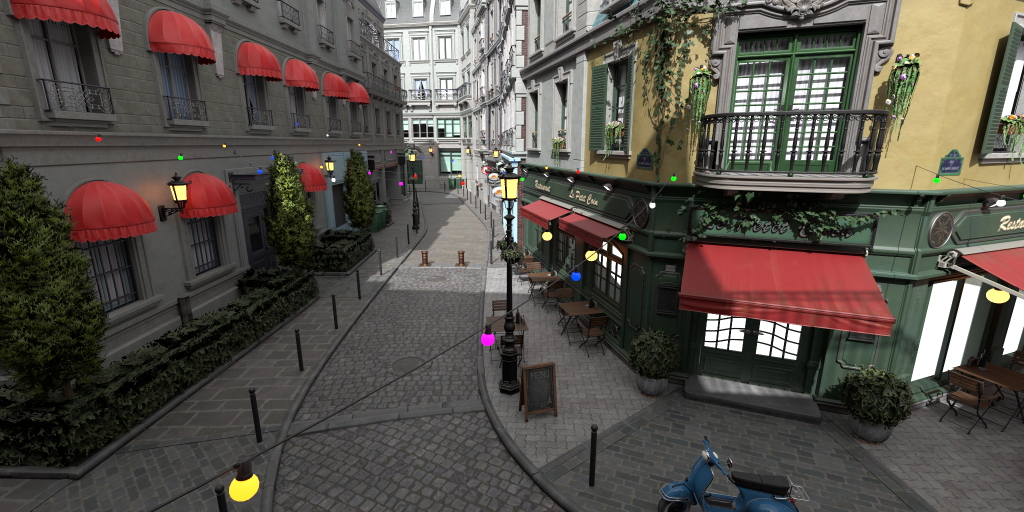}
    \caption{BistroExterior}
\end{subfigure}
\begin{subfigure}{0.32\linewidth}
    \includegraphics[width=\textwidth]{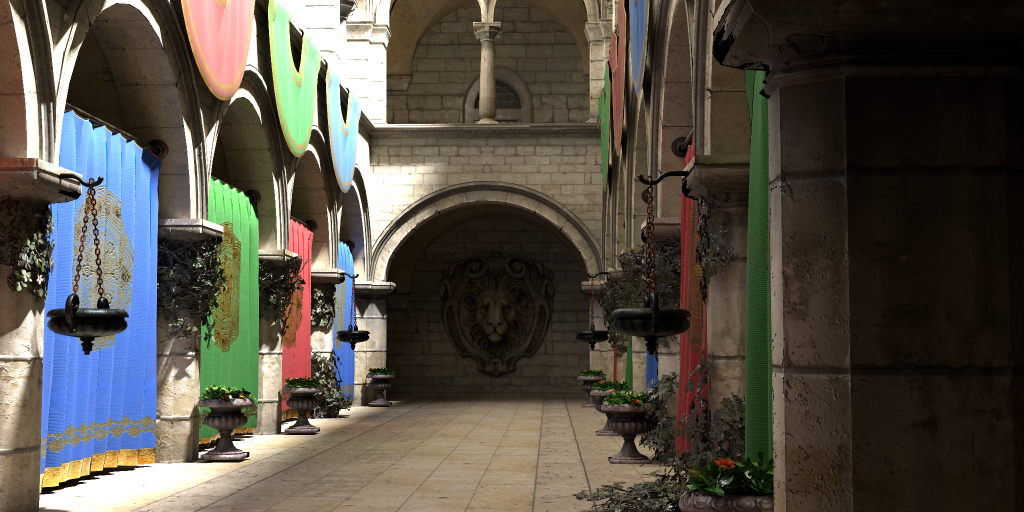}
    \caption{Sponza}
\end{subfigure}
\begin{subfigure}{0.32\linewidth}
    \includegraphics[width=\textwidth]{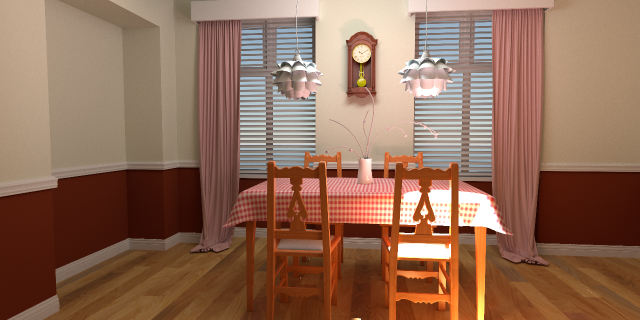}
    \caption{Diningroom}
\end{subfigure}
\begin{subfigure}{0.32\linewidth}
    \includegraphics[width=\textwidth]{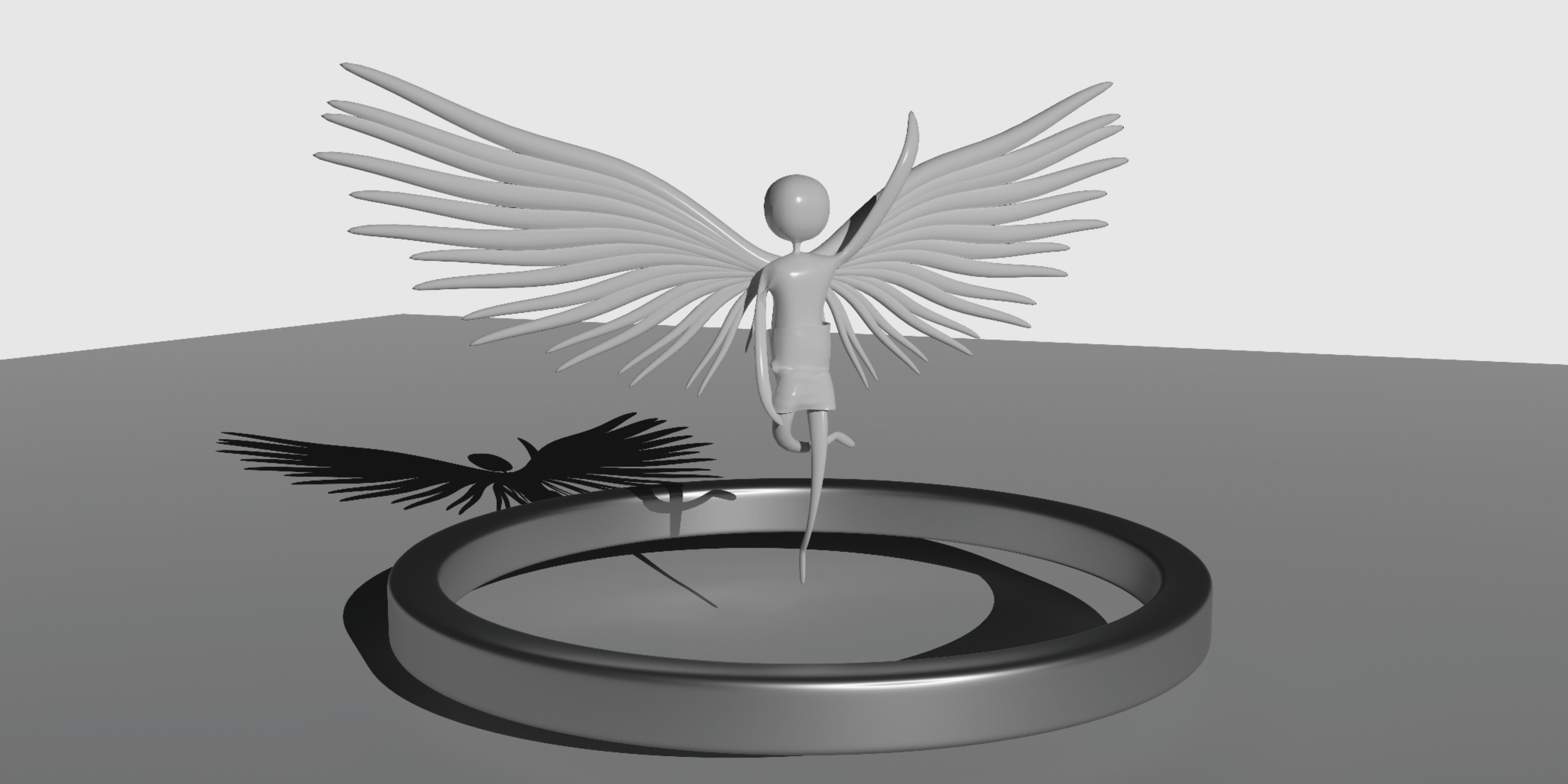}
    \caption{Angel}
\end{subfigure}
\begin{subfigure}{0.32\linewidth}
    \includegraphics[width=\textwidth]{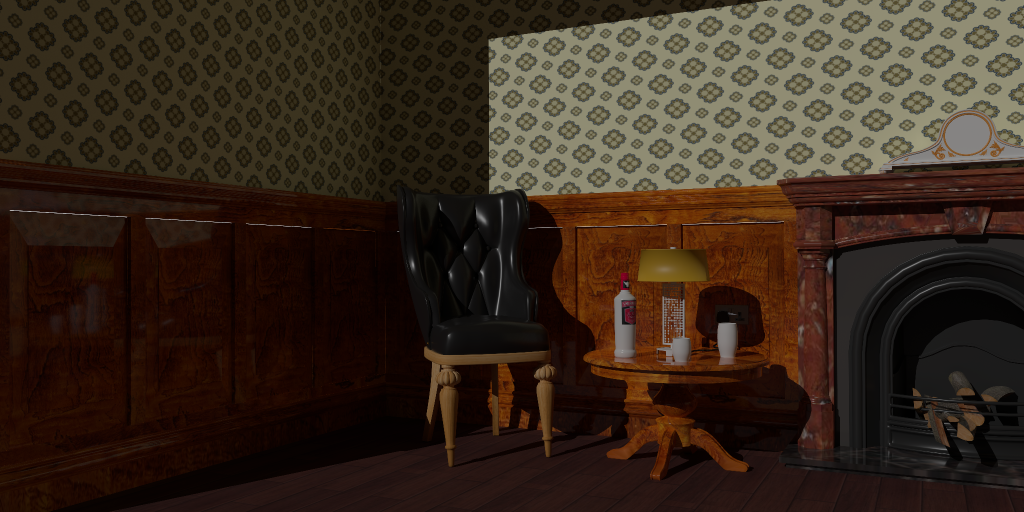}
    \caption{Warmroom}
\end{subfigure}
  \caption{An overview of reference images in our generated dataset.}
  \label{fig:overview_fig}
\end{figure}

\noindent\textbf{Datasets}
As subpixel sampling is a novel strategy, there are no existing datasets available for this purpose. Therefore, we utilized Vulkan\cite{sellers2016vulkan} to construct a hybrid ray tracer, allowing us to generate our subpixel sampling dataset. Our purpose is to optimize our approach for use in 3A gaming and virtual rendering applications. To achieve this, we trained each 3D scene separately rather than training all scenes together, following the same pattern as NVIDIA's DLSS\cite{dlss}. Since our input images were generated at 1/4-spp, a large number of images were required to train a robust denoiser. We trained our method on six distinct scenes, which are shown in \cref{fig:overview_fig}. The BistroInterior and BistroExterior \cite{ORCAAmazonBistro} scenes are complex, containing more than one million triangles and featuring transparency, diffuse, specular, and soft shadow effects. In contrast, the Sponza, Diningroom, Angel, and Warmroom scenes are relatively simple.  All scenes contain 100 to 1000 frames with a resolution of $1024 \times 2048$. We also rendered a validation set of 10 frames and a $50$ frames test set for each scene. The ground truth image is rendered at 32768-spp for reference.

\noindent\textbf{Metrics}
All comparison approaches are evaluated by three image quality metrics, peak signal to noise ratio (PSNR), structural similarity index (SSIM) \cite{SSIM2004}, and root mean squared error (RMSE). Higher PSNR and SSIM imply superior, while lower RMSE indicate better.

\subsection{Implementation Details}

We randomly select $5$ consecutive frames for training each scene. To fully utilize the GPU, we also randomly cropped the inputs, including the noisy image and auxiliary features, to a resolution of 256x256. The kernel size is $3\times3$ at all layers. The weight coefficients for $\mathcal{L}_{s},\ \mathcal{L}_{t},\  \mathcal{L}_{e},\ \mathcal{L}_{w},$  and $\mathcal{L}_{a}$ are 0.7, 0.1, 0.2, 0.4, and 5.0, respectively. We conducted all experiments using the PyTorch framework \cite{Paszke2019} on 8 NVIDIA Tesla A100 GPUs. Adam optimizer\cite{KingmaB2015} with $\beta_1 = 0.9$, $\beta_2 = 0.999$, and $\epsilon=1e-8$ is used with initial learning rate set to $1\times 10^{-4}$. The learning rate is halved at one-third and two-thirds of the total number of iterations. We set batch size to 8 and train our model for 200 epoch. Each scene required approximately 9 hours of training time.

We compare our method to several state-of-the-art Monte Carlo denoising and reconstruction techniques, including the fastest running method RAE \cite{Chaitanya17}, ANF \cite{Mustafa21} which achieves the highest denoising metrics on more than 1-spp images, offline method AFS \cite{Yu2021}, and super-resolution approach NSRR \cite{Xiao2020}. While NSRR is primarily designed for super-resolution, it can also be adapted to fit the sparse sampling task, as detailed in the supplementary materials. Meanwhile, it has practical applications in 3A game rendering, making it a relevant competitor for our study. We re-implemented all methods using their default settings.


\begin{figure*}

\begin{subfigure}{0.35\textwidth}
    \includegraphics[width=\textwidth]{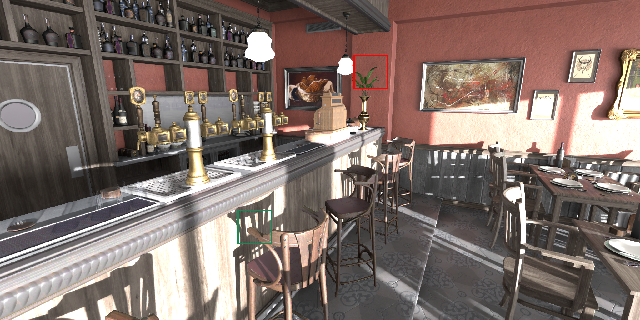}
\end{subfigure}
\begin{subfigure}{0.086\textwidth}
    \includegraphics[width=\textwidth]{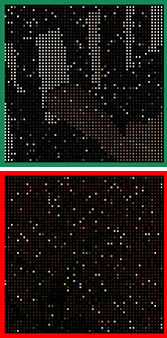}
\end{subfigure}
\begin{subfigure}{0.086\textwidth}
    \includegraphics[width=\textwidth]{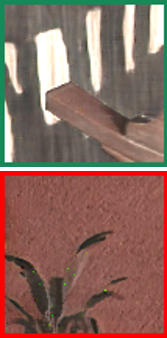}
\end{subfigure}
\begin{subfigure}{0.086\textwidth}
    \includegraphics[width=\textwidth]{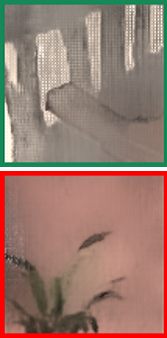}
\end{subfigure}
\begin{subfigure}{0.086\textwidth}
    \includegraphics[width=\textwidth]{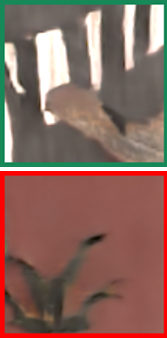}
\end{subfigure}
\begin{subfigure}{0.086\textwidth}
    \includegraphics[width=\textwidth]{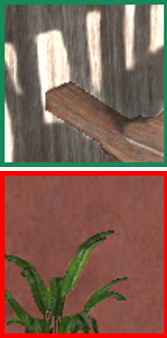}
\end{subfigure}
\begin{subfigure}{0.086\textwidth}
    \includegraphics[width=\textwidth]{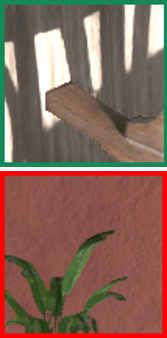}
\end{subfigure}
\begin{subfigure}{0.086\textwidth}
    \includegraphics[width=\textwidth]{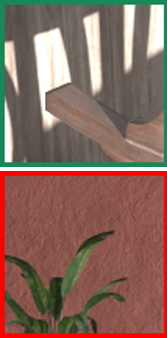}
\end{subfigure}

\begin{subfigure}{0.35\textwidth}
    \includegraphics[width=\textwidth]{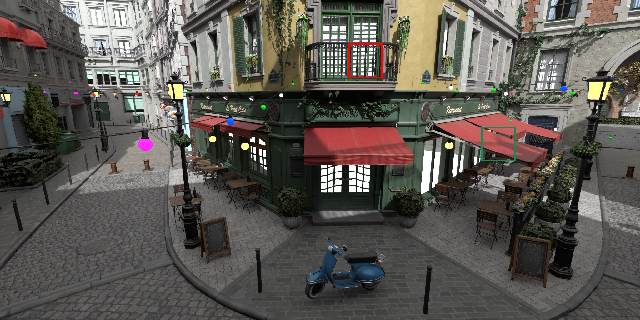}
\end{subfigure}
\begin{subfigure}{0.086\textwidth}
    \includegraphics[width=\textwidth]{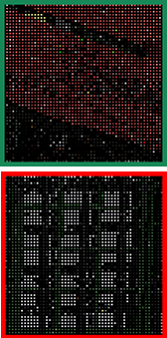}
\end{subfigure}
\begin{subfigure}{0.086\textwidth}
    \includegraphics[width=\textwidth]{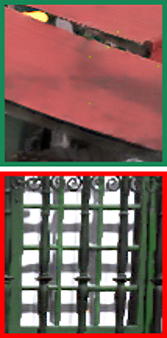}
\end{subfigure}
\begin{subfigure}{0.086\textwidth}
    \includegraphics[width=\textwidth]{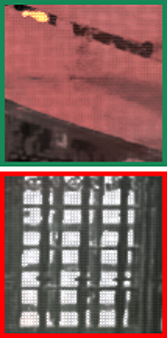}
\end{subfigure}
\begin{subfigure}{0.086\textwidth}
    \includegraphics[width=\textwidth]{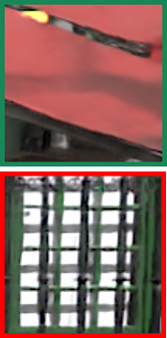}
\end{subfigure}
\begin{subfigure}{0.086\textwidth}
    \includegraphics[width=\textwidth]{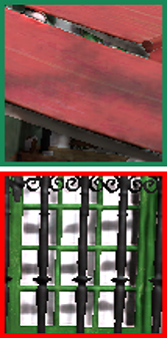}
\end{subfigure}
\begin{subfigure}{0.086\textwidth}
    \includegraphics[width=\textwidth]{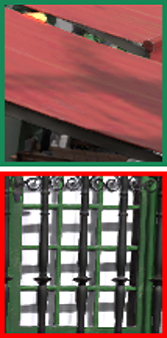}
\end{subfigure}
\begin{subfigure}{0.086\textwidth}
    \includegraphics[width=\textwidth]{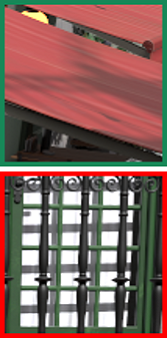}
\end{subfigure}

\begin{subfigure}{0.35\textwidth}
    \includegraphics[width=\textwidth]{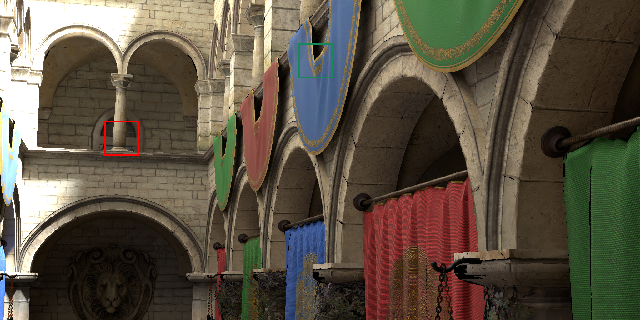}
\end{subfigure}
\begin{subfigure}{0.086\textwidth}
    \includegraphics[width=\textwidth]{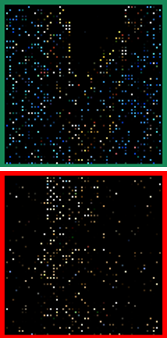}
\end{subfigure}
\begin{subfigure}{0.086\textwidth}
    \includegraphics[width=\textwidth]{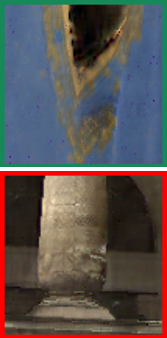}
\end{subfigure}
\begin{subfigure}{0.086\textwidth}
    \includegraphics[width=\textwidth]{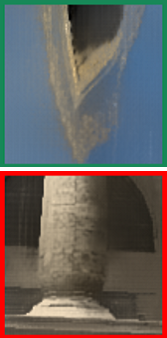}
\end{subfigure}
\begin{subfigure}{0.086\textwidth}
    \includegraphics[width=\textwidth]{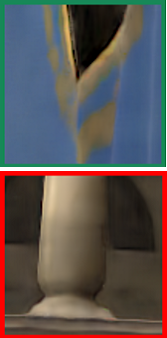}
\end{subfigure}
\begin{subfigure}{0.086\textwidth}
    \includegraphics[width=\textwidth]{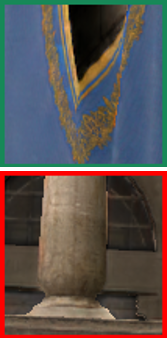}
\end{subfigure}
\begin{subfigure}{0.086\textwidth}
    \includegraphics[width=\textwidth]{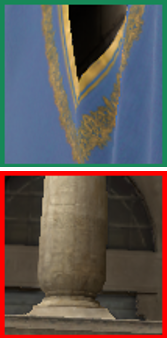}
\end{subfigure}
\begin{subfigure}{0.086\textwidth}
    \includegraphics[width=\textwidth]{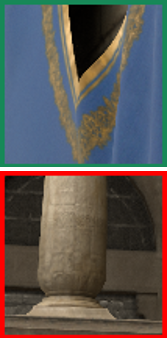}
\end{subfigure}

\begin{subfigure}{0.35\textwidth}
    \includegraphics[width=\textwidth]{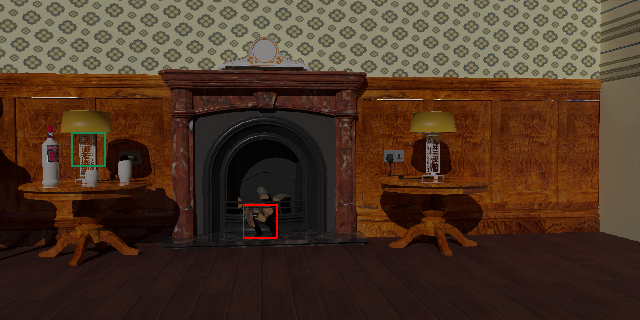}
\end{subfigure}
\begin{subfigure}{0.086\textwidth}
    \includegraphics[width=\textwidth]{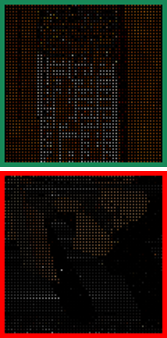}
\end{subfigure}
\begin{subfigure}{0.086\textwidth}
    \includegraphics[width=\textwidth]{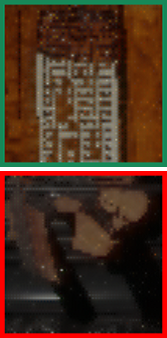}
\end{subfigure}
\begin{subfigure}{0.086\textwidth}
    \includegraphics[width=\textwidth]{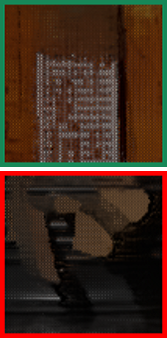}
\end{subfigure}
\begin{subfigure}{0.086\textwidth}
    \includegraphics[width=\textwidth]{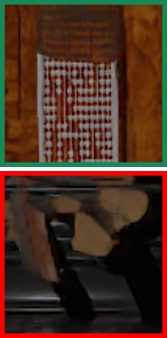}
\end{subfigure}
\begin{subfigure}{0.086\textwidth}
    \includegraphics[width=\textwidth]{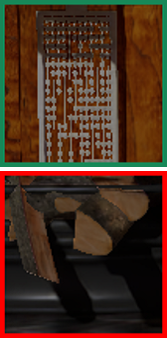}
\end{subfigure}
\begin{subfigure}{0.086\textwidth}
    \includegraphics[width=\textwidth]{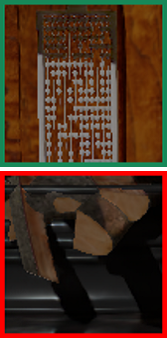}
\end{subfigure}
\begin{subfigure}{0.086\textwidth}
    \includegraphics[width=\textwidth]{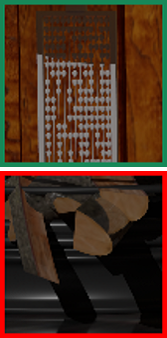}
\end{subfigure}

\begin{subfigure}{0.35\textwidth}
    \includegraphics[width=\textwidth]{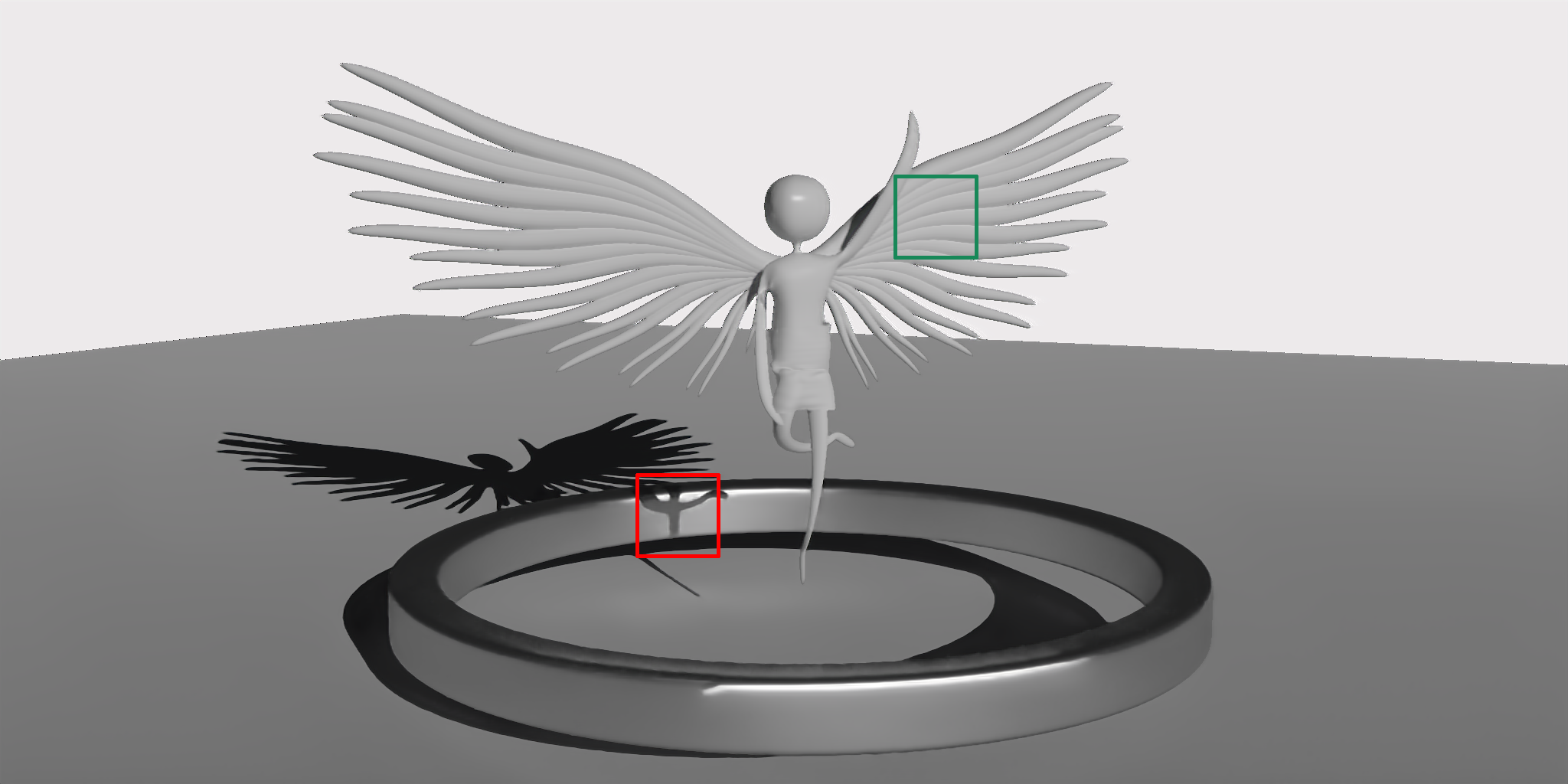}
    \caption{Ours}
\end{subfigure}
\begin{subfigure}{0.086\textwidth}
    \includegraphics[width=\textwidth]{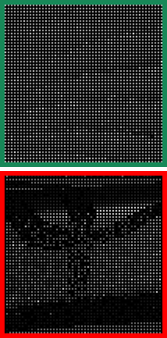}
    \caption{Input}
\end{subfigure}
\begin{subfigure}{0.086\textwidth}
    \includegraphics[width=\textwidth]{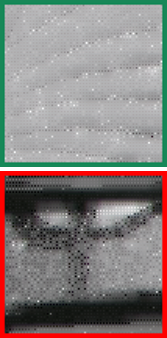}
    \caption{AFS}
\end{subfigure}
\begin{subfigure}{0.086\textwidth}
    \includegraphics[width=\textwidth]{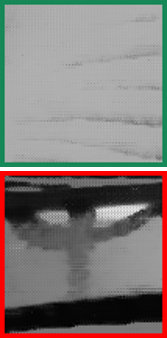}
    \caption{ANF}
\end{subfigure}
\begin{subfigure}{0.086\textwidth}
    \includegraphics[width=\textwidth]{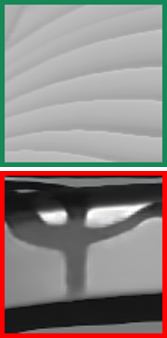}
    \caption{NSRR}
\end{subfigure}
\begin{subfigure}{0.086\textwidth}
    \includegraphics[width=\textwidth]{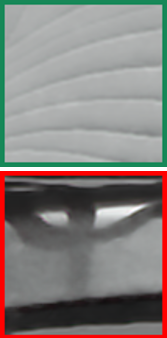}
    \caption{RAE}
\end{subfigure}
\begin{subfigure}{0.086\textwidth}
    \includegraphics[width=\textwidth]{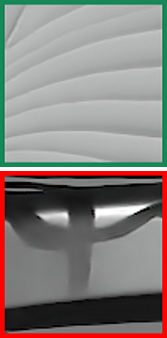}
    \caption{SSR}
\end{subfigure}
\begin{subfigure}{0.086\textwidth}
    \includegraphics[width=\textwidth]{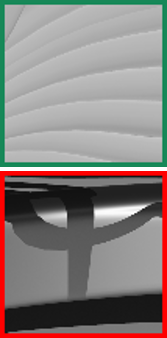}
    \caption{Ref}
\end{subfigure} 

\caption{Visual results on scenes BistroInterior, BistroExterior, Sponza, Warmroom and Angel. }
\label{fig:sup1}
\end{figure*}

\subsection{Time Analysis}

\subsubsection{Rendering}

To showcase the efficiency of our subpixel sampling, we test rendering time of each stage in NVIDIA RTX 3090 GPU at resolution $1024\times2048$, see \cref{table:rendering_time}. The subpixel sampling strategy significantly reduce the sampling time from 12.79ms to 4.35ms, resulting in a 34\% reduction in time costs. With the employment of subpixel sampling, the average total rendering time is 6.92 ms, compared to 15.36 ms without it, resulting in an approximate $3\times$ improvement in rendering cost.

\subsubsection{Reconstruction}

We also conducted an evaluation of the inference time for SSR and compared it against other methods. The comparison was carried out using the same frame for each scene, and the average results are presented in \cref{table:time}, which shows the average inference time for all six scenes at 1024 $\times$ 2048 and 1024 $\times$ 1080 resolution using an NVIDIA Tesla A100. Our SSR is capable of reaching a remarkable 130 FPS when operating at 2K resolution, and 220 FPS at 1080p images. At both resolutions, SSR provides a frame rate improvement of approximately 37\% compared to the previously fastest method.

\cref{table:rendering_time} and \cref{table:time} show the time cost of rendering and reconstruction respectively, while their combined cost is displayed in \cref{fig:tq}, which will be discussed in \cref{sec: qe}.

\begin{table*}[t]
\setlength{\tabcolsep}{1.4mm}{
\begin{tabular}{@{}lcccccccccccccc@{}}
\toprule
\multirow{2}{*}{Method} & \multicolumn{2}{c}{BistroInterior} & \multicolumn{2}{c}{BistroExterior} & \multicolumn{2}{c}{Sponza}      & \multicolumn{2}{c}{Diningroom}  & \multicolumn{2}{c}{Warmroom}    & \multicolumn{2}{c}{Angel}       & \multicolumn{2}{c}{Ave}         \\ \cmidrule(l){2-15} 
   & PSNR             & SSIM            & PSNR             & SSIM            & PSNR           & SSIM           & PSNR           & SSIM           & PSNR           & SSIM           & PSNR           & SSIM           & PSNR           & SSIM           \\ \midrule
AFS\cite{Yu2021}                     & 22.86            & .7650           & 24.60            & .8071           & 25.50          & .8119          & 25.41          & .8637          & 29.55          & .8021          & 22.06          & .8601          & 25.00          & .8183          \\
ANF\cite{Mustafa21}                     & 23.20            & .7583           & 22.14            & .7201           & 23.98          & .8219          & 22.23          & .7226          & 30.91          & .8774          & 25.86          & .8813          & 24.72          & .7969          \\
NSRR\cite{Xiao2020}                    & 23.87            & .8104           & {\ul 25.54}      & {\ul .8538}     & 24.93          & .8113          & 27.17          & .8843          & {\ul 36.40}    & {\ul .9740}    & {\ul 34.94}    & {\ul .9804}    & {\ul 28.81}    & {\ul .8857}    \\
RAE\cite{Chaitanya17}                     & {\ul 24.03}      & {\ul .8351}     & 24.11            & .8006           & {\ul 27.74}    & {\ul .8898}    & {\ul 29.87}    & {\ul .9007}    & 34.32          & .9675          & 29.18          & .9161          & 28.21          & .8849          \\
SSR                     & \textbf{28.99}   & \textbf{.8945}  & \textbf{29.97}   & \textbf{.9121}  & \textbf{31.79} & \textbf{.9410} & \textbf{32.48} & \textbf{.9375} & \textbf{37.34} & \textbf{.9799} & \textbf{38.04} & \textbf{.9876} & \textbf{33.10} & \textbf{.9421} \\ \bottomrule
\end{tabular}
}
\caption{Quantitative comparison results on six scenes. We choose four baseline methods to compare with our SSR method. The best result is in \textbf{bold}, and the {\ul second-best} is underlined in each column.}
\label{table:ssim_compare}
\end{table*}

\begin{table}[t]
\centering
\setlength{\tabcolsep}{1.5mm}{
\begin{tabular}{@{}lcccc@{}}
\toprule
{Strategy} & {R (ms)} & {T\&S (ms)} & {Sampling (ms)} & {Overall(ms)} \\ \midrule
w-SS                      & 0.85                    & 1.72                     & 4.35                          & \textbf{6.92}                         \\
w/o-SS                    & 0.85                    & 1.72                     & 12.79                          & 15.36                        \\ \bottomrule
\end{tabular}
}
\caption{Average rendering time of six scenes. R implies the rendering stage rasterization, and T\&S stands for rendering transparent and shadow stage. w-SS denotes rendering with our subpixel sampling (1/4-spp), while w/o-SS means without it (1-spp).}
\label{table:rendering_time}
\vspace{0.0cm}
\end{table}

\begin{table}[t]
\centering
\setlength{\tabcolsep}{2.6mm}{
\begin{tabular}{lcccc}
\hline
\multirow{2}{*}{Methods} & \multicolumn{2}{c}{1024×2048} & \multicolumn{2}{c}{1024×1080} \\ \cline{2-5} 
                         & Time (ms)        & FPS        & Time (ms)        & FPS        \\ \hline
AFS\cite{Yu2021}                      & 41.8             & 24         & 25.6             & 39         \\
ANF\cite{Mustafa21}                      & 33.0             & 30         & 19.8             & 51         \\
NSRR\cite{Xiao2020}                     & 34.5             & 29         & 21.7             & 46         \\
RAE\cite{Chaitanya17}                      & {\ul 10.4 }            & {\ul 96}         & {\ul 6.22}             & {\ul 160}        \\
SSR                      & \textbf{7.6}              & \textbf{131}        & \textbf{4.56}             & \textbf{220}        \\ \hline
\end{tabular}
}
\caption{Comparison results of inference time. Our SSR can achieve 130 frames per second (FPS) at 2K resolution and 220 FPS at 1080p resolution.}
\label{table:time}
\end{table}

\vspace{0.15cm}
\subsection{Quantitative Evaluation}
\vspace{0.20cm}
\label{sec: qe}
Quantitative comparison results are shown in \cref{table:ssim_compare}. Average results are reported on the 50 test videos of six scenes. Our method delivers the best performance in all scenes. We only show the results of PSNR and SSIM due to space limit, and please refer to our supplemental material for more comparison results. 

\begin{figure}[t]
\begin{center}
 \includegraphics[width=0.5\textwidth]{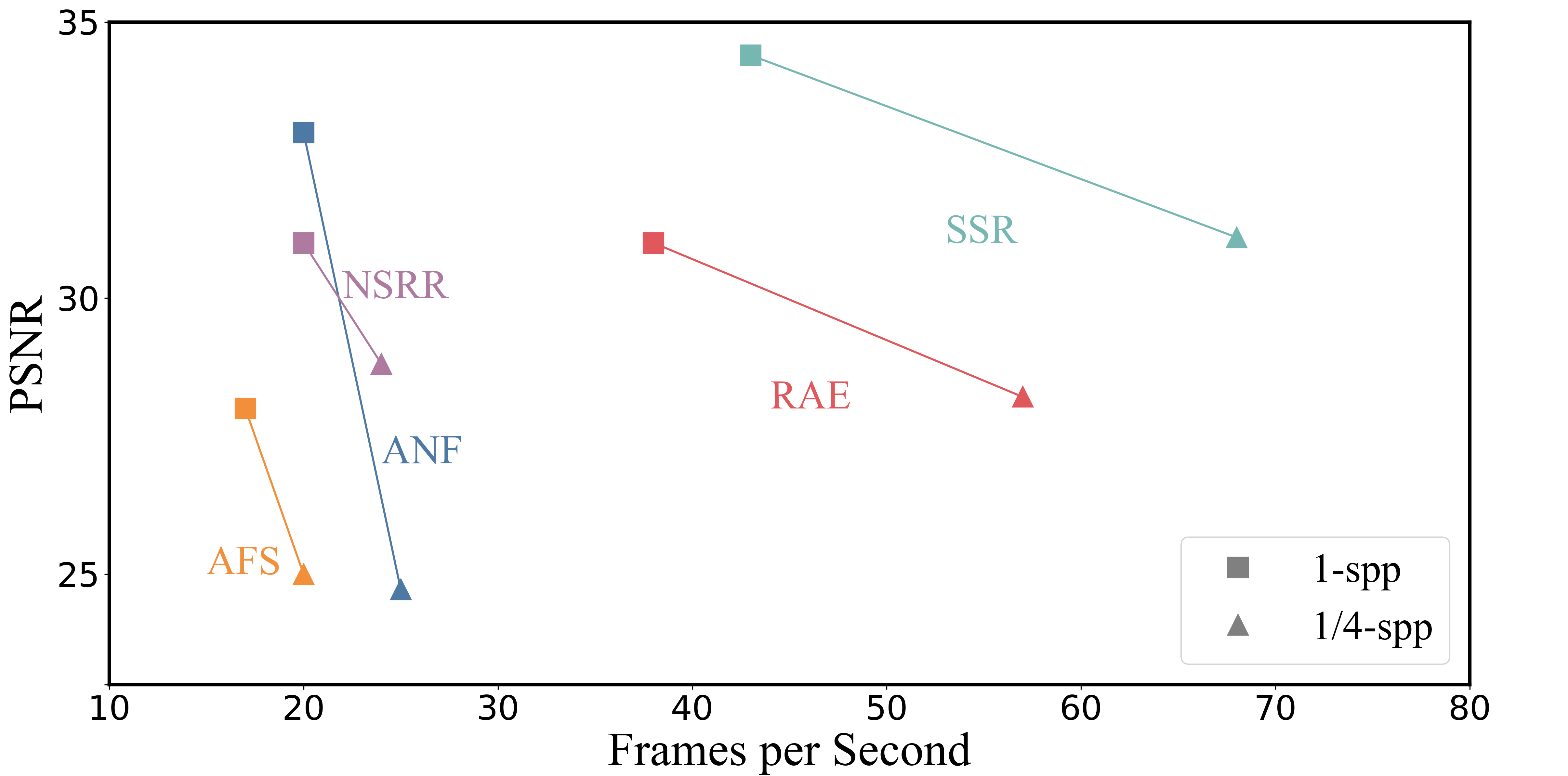}
\end{center}
\caption{ Speed-quality comparison on the 1-spp and 1/4-spp scenes at resolution 1024$\times$2048, where higher PSNR and FPS (top right) is most desirable.}
\label{fig:tq}
\end{figure}

To show the speed and quality improvements of our method more clearly, we generated six scenes without using the subpixel sampling strategy, resulting in 1-spp images with other settings as sane as 1/4-spp scenes in \cref{dataset}. We assessed the entire image generation time, including both rendering and reconstruction costs, and reported the speed and quality comparisons in \cref{fig:tq}. SSR performs well on both 1-spp and 1/4-spp datasets, with tiny declines in quality performance as the sampling rate decreased. (FPS ranging from 43 to 68 and PSNR ranging from 34.40 to 31.10).  In contrast, previous methods aimed at datasets larger than 1-spp exhibited dramatic performance degradation.

\subsection{Qualitative Evaluation}

Here we provide qualitative evaluations of our model. However, to best appreciate the quality of our method, we encourage the reader to watch the supplementary video that contains many example results. \cref{fig:sup1} compares reconstructed images in several scenes visually. We included all comparison results for six scenes in the supplementary material. Our method outperforms all other methods by a considerable margin across all scenes. Previous state-of-the-art methods, designed for denoising renderings with more than 1-spp, are not as effective at denoising renderings at 1/4-spp. AFS was originally designed for offline rendering, and transformer models \cite{vaswani2017attention,SwinTransformer2021} require significant memory to train and perform inference. RAE, NSRR, and ANF feed previous and current features directly into the network, which leads to blurred and aliased details. Different from them, SSR computes the correlation for each pixel between normal and depth features of the current and previous frame, thus having the capacity to generate high-quality details.

\begin{figure}[t]
  \centering
\begin{subfigure}{0.112\textwidth}
    \includegraphics[width=\textwidth]{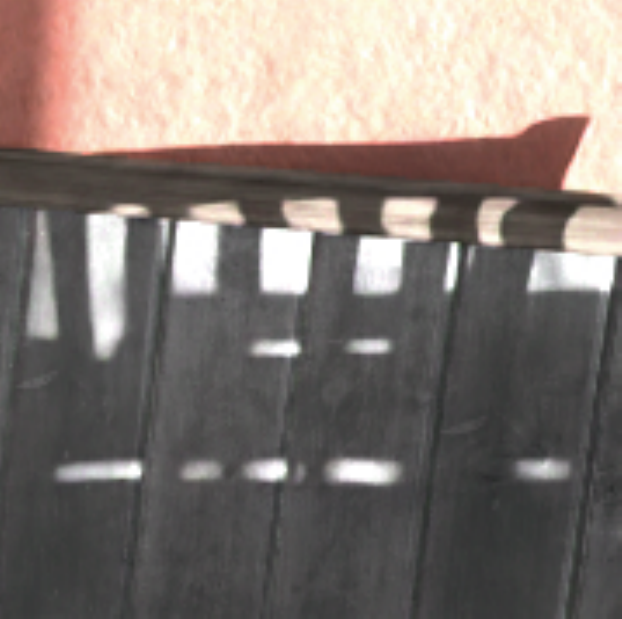}
    \caption{w/o-shadow}
\end{subfigure}
\begin{subfigure}{0.112\textwidth}
    \includegraphics[width=\textwidth]{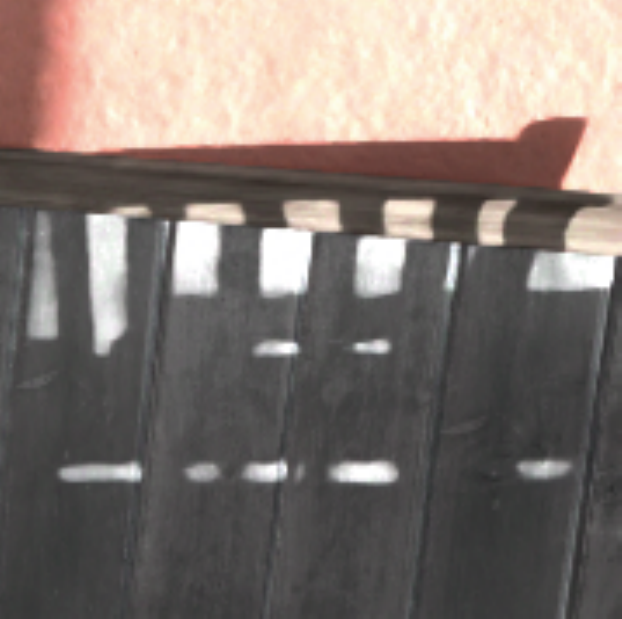}
    \caption{w-shadow}
\end{subfigure}
\begin{subfigure}{0.112\textwidth}
    \includegraphics[width=\textwidth]{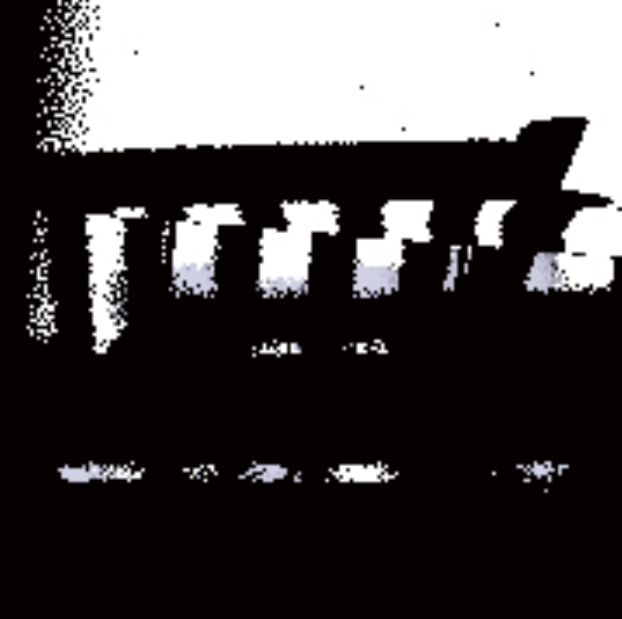}
    \caption{Noisy}
\end{subfigure}
\begin{subfigure}{0.112\textwidth}
    \includegraphics[width=\textwidth]{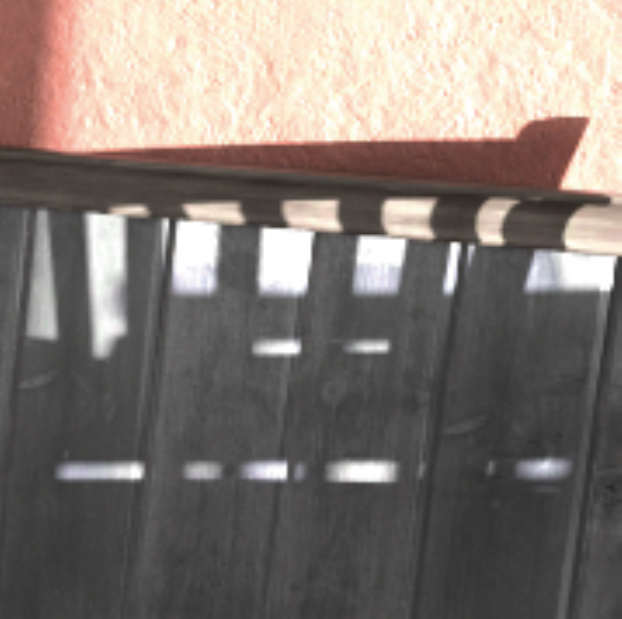}
    \caption{Ref}
\end{subfigure}

  \caption{ (a) and (b) show the reconstruction results without and with
  employing shadow(c), respectively. (d) is the 32768-spp reference image. The shadow feature assists SSR in pinpointing more precise contours. }
  \label{fig:shadow}
\end{figure}

\begin{table}[t]
\centering
\setlength{\tabcolsep}{3.0mm}{
\begin{tabular}{@{}l|l|l|l|l@{}}
\toprule
Method      & RN & TFA & WP & PSNR/SSIM   \\ \midrule
Base        & \Checkmark  &     &    & 29.55/.8952 \\
Base+TFA    & \Checkmark  & \Checkmark   &    & {\ul 32.13}/{\ul .9245} \\
Base+TFA+WP & \Checkmark  & \Checkmark   & \Checkmark  & \textbf{33.10}/\textbf{.9421} \\ \bottomrule
\end{tabular}

}
\caption{Ablation study. We evaluate different modules
on six scenes. PSNR and SSIM are shown on average.}
\label{table:network_module}
\end{table}

\subsection{Ablation Study}
\subsubsection{GBuffers ablation}

We incorporated certain features from the Gbuffers that have not been utilized in existing Monte Carlo denoising methods, and conducted corresponding ablation experiments to investigate their effectiveness.

\noindent\textbf{Shadow.}
Our training images are generated by subpixel sampling. As a result of 1/4-spp light occlusion, more than three-quarters of the pixels remain at a value of zero, which motivate us to identify reliable pixels to train our model. Thus, we take the shadow feature as an additional input. Our feature accumulator collects the noisy shadows from the current frame and combines them with the history shadow. This accumulated shadow information aids in detecting continuous edges of shadows and improve temporal stability, as shown in \cref{fig:shadow}. 

\begin{figure}[t]
  \centering
\begin{subfigure}{0.112\textwidth}
    \includegraphics[width=\textwidth]{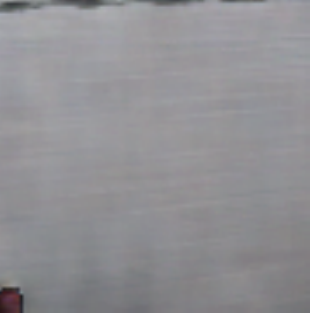}
    \caption{w/o-T}
\end{subfigure}
\begin{subfigure}{0.112\textwidth}
    \includegraphics[width=\textwidth]{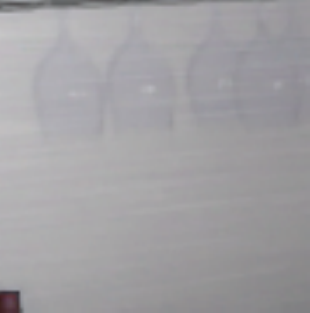}
    \caption{w-T}
\end{subfigure}
\begin{subfigure}{0.112\textwidth}
    \includegraphics[width=\textwidth]{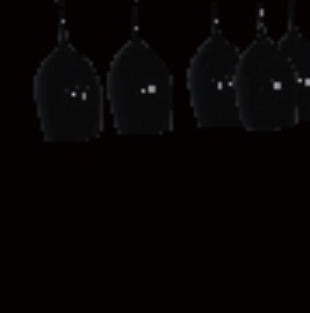}
    \caption{Transparent}
\end{subfigure}
\begin{subfigure}{0.112\textwidth}
    \includegraphics[width=\textwidth]{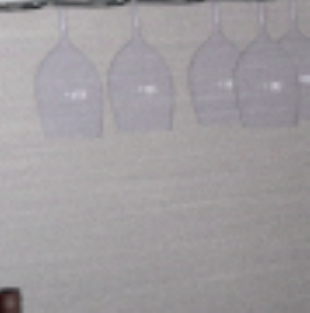}
    \caption{Ref}
\end{subfigure}

  \caption{ (a) and (b) show the reconstruction results without and with
  employing transparent (c), respectively. (d) is the 32768-spp reference image. SSR can capture the information from the transparent features and restore clear transparent objects.}
  \label{fig:trans}
\end{figure}

\noindent\textbf{Transparent.}
 We also append the transparent feature into SSR for training, but we do not accumulate transparent before feeding it into the reconstruction network. This choice is made since the transparent feature is scarce in a whole image, and contains rare noise, as shown in \cref{gb4}. Accumulating the transparent feature yields minor improvement, but also comes with an increased time cost. So we choose to feed the transparent feature into our reconstruction network directly. With utilizing transparent, SSR acquires the ability of producing transparent objects, such as clear glass cups, as illustrated in \cref{fig:trans}. Additionally, in cases where a scene does not contain any transparent objects, such as the BistroExterior scene, we still include the transparent feature with a value of zero.

Without using shadow and transparency, SSR only achieves a PSNR of 27.67 when testing on BistroInterior, while employing shadow brings an improvement to 28.22. By including both shadow and transparency, our model produces a higher PSNR of 28.99.

\subsubsection{Network ablation}
\label{sec:ab}
We verify the effectiveness of different modules in our approach, including the temporal feature accumulator (TFA) and the warped previous output (WP), as shown in \cref{table:network_module}. Results are reported on average of six scenes. Without TFA, the PSNR decreased by 2.58 compared to the results with TFA, and the temporal stability in the video results decreased obviously. Please see supplemental video for more details. The utilizing of WP also show its efficacy in the third row.

\vspace{0.15cm}
\section{Conclusion}

We presented a novel Monte Carlo sampling strategy called subpixel sampling to enable faster rendering. A denoising network subpixel sampling reconstruction (SSR) was also introduced to recover high-quality image sequences at real-time frame rates from rendering results using subpixel sampling pattern. Experiments showed that our method produces superior denoised results compared to existing state-of-the-art approaches and achieves real-time performance at 2K resolution.

\noindent\textbf{Limitations and Future Work.}
While our method provides a real-time pattern for reconstructing high-quality images from subpixel sampling, the inference time still has room for improvement. 16-bit precision TensorRT will be used for acceleration and deploying SSR on our game engine in the future. Additionally, we explored the integration of Swin Transformer \cite{SwinTransformer2021} into the first layer of our reconstruction network, which improved PSNR by approximately 0.23 but increased inference time by 1.1 ms. Speed-quality trade-off is a critical goal in our future work.

{\small
\bibliographystyle{ieee_fullname}
\bibliography{egbib}
}

\end{document}